\documentclass{article}

\usepackage{graphicx}
\usepackage{arxiv}
\usepackage{color}
\usepackage{amsthm}
\usepackage[utf8]{inputenc} 
\usepackage[T1]{fontenc}    
\usepackage{lineno,hyperref}       
\usepackage{url}            
\usepackage{booktabs}       
\usepackage{amsfonts}       
\usepackage{nicefrac}       
\usepackage{microtype}      
\usepackage{lipsum}
\usepackage{amssymb}
\usepackage{amsmath}
\usepackage{bm}
\usepackage{multirow}
\usepackage{diagbox}
\usepackage{authblk}
\usepackage[dvipsnames]{xcolor}
\usepackage{float}
\usepackage{xcolor}
\usepackage{tikz-cd}
\usepackage{algorithm}
\usepackage{algpseudocode}
\newtheorem{definition}{Definition}[section]
\newtheorem{proposition}{Proposition}[section] 
\newtheorem{lemma}{Lemma}[section] 
\newtheorem{theorem}{Theorem}[section] 
\newtheorem{problem}{Problem}
\newtheorem*{assumption*}{\assumptionnumber}
\providecommand{\assumptionnumber}{}
\makeatletter
\def\algbackskip{\hskip\dimexpr-\algorithmicindent+\labelsep}
\def\LState{\State \algbackskip}%
\def\R{{\mathbb R}}
\def\N{{\mathbb N}}
\def\E{{\mathbb E}}
\def\curlyA{{\mathcal A}}

\def\curlyF{{\mathcal F}}

\def\curlyC{{\mathcal C}}
\def\curlyD{{\mathcal D}}
\def\curlyG{{\mathcal G}}
\def\curlyH{{\mathcal H}}
\def\curlyE{{\mathcal E}}

\def\curlyL{{\mathcal L}}

\def\curlyN{{\mathcal N}}

\def\curlyU{{\mathcal U}}
\def\curlyX{{\mathcal X}}
\def\curlyY{{\mathcal Y}}

\makeatother
\usepackage{nomencl}
\makenomenclature

\makeatletter
\newcommand{\printfnsymbol}[1]{%
	\textsuperscript{\@fnsymbol{#1}}%
}
\makeatother

\title{Learning Operators with Coupled Attention}

\begin{document}

\author[1]{Georgios Kissas \thanks{These authors contributed equally.}  }
\author[2]{Jacob Seidman \printfnsymbol{1}}
\author[2]{Leonardo Ferreira Guilhoto}
\author[3]{\\ Victor M. Preciado} 
\author[3]{George J. Pappas}
\author[1]{Paris Perdikaris}
\affil[1]{Department of Mechanical Engineering 
	and Applied Mechanics\\
	University of Pennsylvania\\
	Philadelphia, PA 19104 }
\affil[2]{Graduate Group in Applied Mathematics and Computational Science \\
	University of Pennsylvania\\
	Philadelphia, PA 19104 }
\affil[3]{	Department of Electrical and Systems Engineering \\
	University of Pennsylvania\\
	Philadelphia, PA 19104 }
		
\maketitle

\begin{abstract}
Supervised operator learning is an emerging machine learning paradigm with applications to modeling the evolution of spatio-temporal dynamical systems and approximating general black-box relationships between functional data. We propose a novel operator learning method, LOCA (Learning Operators with Coupled Attention), motivated from the recent success of the attention mechanism. In our architecture, the input functions are mapped to a finite set of features which are then averaged with attention weights that depend on the output query locations. By coupling these attention weights together with an integral transform, LOCA is able to explicitly learn correlations in the target output functions, enabling us to approximate nonlinear operators even when the number of output function in the training set measurements is very small. Our formulation is accompanied by rigorous approximation theoretic guarantees on the universal expressiveness of the proposed model. Empirically, we evaluate the performance of LOCA on several operator learning scenarios involving systems governed by ordinary and partial differential equations, as well as a black-box climate prediction problem. Through these scenarios we demonstrate state of the art accuracy, robustness with respect to noisy input data, and a consistently small spread of errors over testing data sets, even for out-of-distribution prediction tasks.
\end{abstract}


\section{Introduction}

The great success of modern deep learning lies in its ability to approximate maps between finite-dimensional vector spaces, as in computer vision \cite{santhanam2017generalized}, natural language processing \cite{vaswani2017attention}, precision medicine \cite{rajkomar2019machine}, bio-engineering \cite{kissas2020machine}, and other data driven applications. A  particularly successful class of such models are those built with the attention mechanism \cite{Bahdanau2015}. For example, the Transformer is an attention-based architecture that has recently produced state of the art performance in natural language processing \cite{vaswani2017attention}, computer vision \cite{dosovitskiy2020image,parmar2018image}, and audio signal analysis \cite{gong2021ast,huang2018music}.

Another active area of research is applying machine learning techniques to approximate operators between spaces of functions. These methods are particularly attractive for many problems in computational physics and engineering where the goal is to learn the functional response of a system from a functional input, such as an initial/boundary condition or forcing term. In the context of learning the response of systems governed by differential equations, these learned models can function as fast surrogates of traditional numerical solvers.

\begin{figure}[ht]
\centering
\includegraphics[width=.9\textwidth]{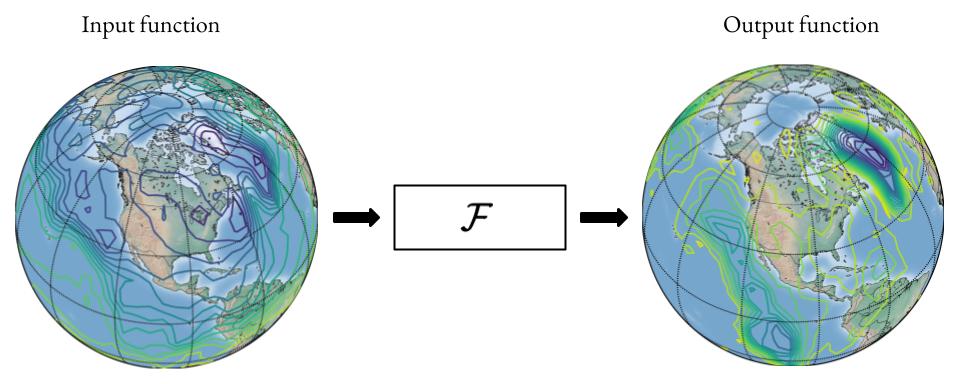}
\caption{ An example sketch of operator learning for climate modeling: By solving an operator learning problem, we can approximate an infinite-dimensional map between two functions of interest, and then predict one function using the other. For example, by providing the model with an input function, e.g. a surface air temperature field, we can predict an output function, e.g. the corresponding surface air pressure field.}
\label{fig:op_sketch}
\end{figure}

For example, in climate modelling one might wish to predict the pressure field over the earth from measurements of the surface air temperature field. The goal is then to learn an operator, $\mathcal{F}$, between the space of temperature functions to the space of pressure functions (see Figure \ref{fig:op_sketch}). An initial attempt at solving this problem might be to take a regular grid of measurements over the earth for the input and output fields and formulate the problem as a (finite-dimensional) image to image regression task. While architectures such as convolutional neural networks may perform well under this setting, this approach can be somewhat limited. For instance, if we desired the value of the output at a query location outside of the training grid, an entirely new model would need to be built and tuned from scratch. This is a consequence of choosing to discretize the regression problem before building a model to solve it. If instead we formulate the problem and model at the level of the (infinite-dimensional) input and output function spaces, and \emph{then} make a choice of discretization, we can obtain methods that are more flexible with respect to the locations of the point-wise measurements.

Formulating models with functional data is the topic of Functional Data Analysis (FDA) \cite{ramsay1982data, ramsay1991some}, where parametric, semi-parametric or non-parametric methods operate on functions in infinite-dimensional vector spaces. A useful class of non-parametric approaches are Operator-Valued Kernel methods. These methods generalize the use of scalar-valued kernels for learning functions in a Reproducing Kernel Hilbert Space (RKHS) \cite{hastie2009elements} to RKHS's of operators. Kernel methods were thoroughly studied in the past \cite{hofmann2008kernel, shawe2004kernel} and have been successfully applied to nonlinear and high-dimensional problem settings \cite{takeda2007kernel, dou2020training}. Previous work has successfully extended this framework to learning operators between more general vector spaces as well \cite{micchelli2005learning,caponnetto2008universal,kadri2010nonlinear,kadri2016operator, owhadi2020ideas}. This framework is particularly powerful as the inputs can be continuous or discrete, and the underlying vector spaces are typically only required to be normed and separable.

A parametric-based approach to operator learning was introduced in Chen \textit{et. al.} \cite{chen1995universal} where the authors proposed a method for learning non-linear operators based on a one-layer feed-forward neural network architecture. Moreover, the authors presented a universal approximation theorem which ensures that their architecture can approximate any continuous operator with arbitrary accuracy. Lu \textit{et. al.} \cite{lu2019deeponet} gave an extension of this architecture, called DeepONet, built with multiple layer feed-forward neural networks, and demonstrated effectiveness in approximating the solution operators of various differential equations. In follow up work, error estimates were derived for some specific problem scenarios \cite{lanthaler2021error}, and several applications have been pursued \cite{cai2020deepm, di2021deeponet, lin2021operator}. An extension of the DeepONet was proposed by Wang et. al. \cite{wang2021learning,wang2021long, wang2021improved}, where a regularization term is added to the loss function to enforce known physical constraints, enabling one to predict solutions of parametric differential equations, even in the absence of paired input-output training data.

Another parametric approach to operator learning is the Graph Neural Operator proposed by Li \textit{et. al.} \cite{li2020neural}, motivated by the solution form of linear partial differential equations (PDEs) and their Greens' functions. As an extension of this work, the authors also proposed a Graph Neural Operator architecture where a multi-pole method is used sample the spatial grid \cite{li2020multipole} allowing the kernel to learn in a non-local manner. In later published work, this framework has been extended to the case where the integral kernel is stationary, enabling one to efficiently compute the integral operator in the Fourier domain \cite{li2020fourier}.

Both the Fourier Neural Operator and the DeepONet methods come with theoretical guarantees of universal approximation, meaning that under some assumptions these classes of models can approximate any continuous operator to arbitrary accuracy. Other parametric-based models include a deep learning approach for directly approximating the Green's function of differential equations \cite{gin2021deepgreen}, a multi-wavelet approach for learning projections of an integral kernel operator to approximate the true operator and a random feature approach for learning the solution map of PDEs \cite{nelsen2020random}, but no theoretical guarantees of the approximation power of these approaches are presented.

While some of the previously described operator learning methods can be seen as generalizations of deep learning architectures such as feed-forward and convolutional neural networks, in this paper we are motivated by the success of the attention mechanism to propose a new operator learning framework. Specifically, we draw inspiration from the Badhanau attention mechanism \cite{Bahdanau2015}, which first constructs a feature representation of the input and then averages these features with a distribution that depends on the argument of the output function to obtain its value. We will also use the connection between the attention mechanism and kernel methods \cite{tsai2019transformer} to couple these distributions together in what we call a {\em Kernel-Coupled Attention} mechanism. This will allow our framework to explicitly model correlations within the output functions of the operator. Moreover, we prove that under certain assumptions the model satisfies a universal approximation property.

The main contributions of this work can be summarized in the following points:
\begin{itemize}

\item  \textbf{Novel Architecture:} We propose an operator learning framework inspired by the attention mechanism, operator approximation theory, and the Reproducing Kernel Hilbert Space (RKHS) literature. To this end, we introduce a novel {\em Kernel-Coupled Attention} mechanism to explicitly model correlations between the output functions' query locations.

\item  \textbf{Theoretical Guarantees:} We prove that the proposed framework satisfies a universal approximation property, that is, it can approximate any continuous operator with arbitrary accuracy.

\item \textbf{Data Efficiency:} By modelling correlations between output queries, our model can achieve high performance when trained with only a small fraction (6-12\%) of the total available labeled data compared to competing methods.

\item  \textbf{Robustness:} Compared to existing methods, our model demonstrates superior robustness with respect to noise corruption in the training and testing inputs, as well as randomness in the model initialization. Our model's performance is stable in that the errors on the test data set are consistently concentrated around the median with significantly fewer outliers compared to other methods.

\item \textbf{Generalization:} On a real data set of Earth surface air temperature and pressure measurements, our model is able to learn the functional relation between the two fields with high accuracy and extrapolate beyond the training data. On synthetic data we demonstrate that our model is able to generalize better than competing methods over increasingly out-of-distribution examples.

\end{itemize}

The paper is structured as follows. In Section \ref{sec:problem_formulation} we introduce the supervised operator learning problem. In Section \ref{sec:proposed_model}, we introduce the general form of the model and in following subsections present the construction of its different components. In Section \ref{sec: univ} we prove theoretical results on the approximation power of this class of models. In Section \ref{sec:implementation of method} we present the specific architecture choices made for implementing our method in practice. Section \ref{sec: related methods} discusses the similarities and differences of our model with related operator learning approaches. In Section \ref{sec:Results}, we demonstrate the performance of the proposed methodology across different benchmarks in comparison to other state-of-the-art methods. In Section \ref{sec:discussion}, we discuss our main findings, outline potential drawbacks of the proposed method, and highlight future directions emerging from this study. 

\section{Problem Formulation}\label{sec:problem_formulation}

We now provide a formal definition of the operator learning problem. Given $\curlyX \subset \R^{d_x}$, $\curlyY \subset \R^{d_y}$, we will refer to a point $x \in \curlyX$ as an \emph{input location} and a point $y \in \curlyY$ as a \emph{query location}. Denote by $C(\curlyX, \R^{d_u})$ and $C(\curlyY, \R^{d_s})$ the spaces of continuous functions from $\curlyX \to \R^{d_u}$ and $\curlyY \to \R^{d_s}$, respectively. We will refer to $C(\curlyX, \R^{d_u})$ as the space of \emph{input functions} and $C(\curlyY, \R^{d_s})$ the space of \emph{output functions}. For example, in Figure \ref{fig:op_sketch}, if we aim to learn the correspondence between a temperature field over the earth and the corresponding pressure field, $u \in C(\curlyX, \R)$ would represent the temperature field and $s \in C(\curlyY, \R)$ would be a pressure field, where $\curlyX = \curlyY$ represents the surface of the earth. With a data set of of input/output function pairs, we formulate the supervised operator learning problem as follows.
\begin{problem} \label{main prob} 
Given $N$ pairs of input and output functions $\{ u^\ell(x), s^\ell(y)\}_{\ell=1}^N$ generated by some possibly unknown ground truth operator $\curlyG: C(\mathcal{X}, \R^{d_u}) \to C(\curlyY, \R^{d_s})$ with $u^\ell \in C(\mathcal{X}, \R^{d_u})$ and $s^\ell \in C(\curlyY, \R^{d_s})$, learn an operator $\curlyF: C(\mathcal{X}, \R^{d_u}) \to C(\curlyY, \R^{d_s})$, such that for $\ell = 1,\ldots, N$,
\begin{align*}
    \curlyF(u^\ell) = s^\ell.
\end{align*}
\end{problem}

This problem also encompasses scenarios where more structure is known about the input/output functional relation. For example, $u$ could represent the initial condition to a partial  differential equation and $s$ the corresponding solution. In this case, $\curlyG$ would correspond to the true solution operator and $\curlyF$ would be an approximate surrogate model. Similarly, $u$ could represent a forcing term in a dynamical system described by an ordinary differential equation, and $s$ the resulting integrated trajectory. 
In these two scenarios there do exist a suite of alternate methods to obtain the solution function $s$ from the input $u$, but with an appropriate choice of architecture for $\curlyF$ the approximate model can result in significant computational speedups and the ability to efficiently compute sensitivities with respect to the inputs using tools like automatic differentiation.

Note that while the domains $\curlyX$ and $\curlyY$ need not be discrete sets, in practice we may only have access to the functions $u^\ell$ and $s^\ell$ evaluated at finitely many locations. However, we take the perspective that it is beneficial to formulate the model with continuously sampled input data, and consider the consequences of discretization at implementation time. As we shall see, this approach will allow us to construct a model that is able to learn operators over multiple output resolutions simultaneously.

\section{Proposed Model: Learning Operators with Coupled Attention (LOCA)}\label{sec:proposed_model}

We will construct our model through the following two steps. Inspired by the attention mechanism \cite{Bahdanau2015}, we will first define a class of models where the input functions $u$ are lifted to a feature vector $v(u) \in \R^{n \times d_s}$. Each output location $y\in\curlyY$ will define $d_s$ probability distributions $\varphi(y) \in \prod_{i=1}^{d_s}\Delta^n$, where $\Delta^n$ is the the $n$-simplex. The forward pass of the model is then computed by averaging the rows of $v(u)$ over the probability distributions $\varphi(y)$. 

Next, we augment this model by coupling the probability distributions $\varphi(y)$ across different query points $y \in \curlyY$. This is done by acting on a proposal score function $g: \curlyY \to \R^{n \times d_s}$ with a kernel integral operator. The form of the kernel determines the similarities between the resulting distributions. We empirically demonstrate that the coupled version of our model is more accurate compared to the uncoupled version when the number of output function evaluations per example is small.

\subsection{The Attention Mechanism}

The attention mechanism was first formulated in Bahdanau \textit{et. al.} \cite{Bahdanau2015} for use in language translation. The goal of their work was to translate an input sentence in a given language $\{u_1, \ldots, u_{T_u}\}$ to a sentence in another language $\{s_1, \ldots, s_{T_s}\}$. A context vector $c_i$ was associated to each index of the output sentence, $i \in \{1,\dots,T_s\}$, and used to construct a probability distribution of the $i$-th word in the translated sentence, $s_i$. The attention mechanism is a way to construct these context vectors by averaging over features associated with the input in a way that depends on the output index $i$.
 
More concretely, the input sentence is first mapped to a collection of features $\{v_1, \ldots, v_{T_u}\}$. Next, depending on the input sentence and the location/index $i$ in the output (translated) sentence, a discrete probability distribution $\{\varphi_{i1}, \ldots, \varphi_{iT_u}\}$ is formed over the input indices such that
$$\varphi_{ij} \geq 0, \quad \sum_{j=1}^{T_u} \varphi_{ij} = 1. $$
The context vector at index $i$ is then computed as
\begin{align*} \label{eq: generic attn}
c_i = \sum_{j=1}^{T_u}\varphi_{ij}v_j.
\end{align*}
If the words in the input sentence are represented by vectors in $\R^d$, and the associated features and context vector are in $\R^l$, the attention mechanism can be represented by the following diagram.
\begin{equation*}
\begin{tikzcd}
\mathcal  [T_s] \times \R^{T_u \times d}   \arrow[r, "\mathrm{Attn}"] \arrow[d, "{(\varphi, v)}", swap] & \mathbb R^l \\
 \Delta^{T_u} \times  \R^{T_u \times l} \arrow[ru, "\mathbb E", swap]&
\end{tikzcd}
\end{equation*}

We will apply this attention mechanism to learn operators between function spaces by mapping an input function $u$ to a finite set of features $v(u) \in \R^{n \times d_s}$, and taking an average over these features with respect to $d_s$ distributions $\varphi(y) \in \prod_{k=1}^{d_s} \Delta^n$ that depend on the query location $y \in \curlyY$ for the output function. That is,
\begin{equation*}
    \curlyF(u)(y) := \E_{\varphi(y)}[v(u)],
\end{equation*}
where $v(u) \in \mathbb{R}^{n \times d_s}$, $\varphi$ is a function from $y \in \curlyY$ to $d_s$ copies of the $n$-dimensional simplex $\Delta^{n}$, and $\mathbb E: \prod_{k=1}^{d_s} \Delta^n \times \R^{n \times d_s} \to \R^{d_s}$ is an expectation operator that takes $(\varphi, v) \mapsto \sum_i \varphi_i \odot v_i$, where $\odot$ denotes an element-wise product. This can be represented by the following diagram.
\begin{equation*}
\begin{tikzcd}
\mathcal \curlyY \times C(\curlyX,\R^{d_u})  \arrow[r, "\mathcal F"] \arrow[d, "{(\varphi, v)}", swap] & \mathbb R^{d_s} \\
\mathbb \prod_{k=1}^{d_s}\Delta^n \times \R^{n\times d_s}  \arrow[ru, "\mathbb E", swap]&
\end{tikzcd}
\end{equation*}

In the next section, we will construct the function $\varphi$ and provide a mechanism for enabling the coupling of its values across varying query locations $y \in \curlyY$. Later on, we will see that this allows the model to perform well even when trained on small numbers of output function measurements per input function.

\subsection{Kernel-Coupled Attention Weights}

In order to model correlations among the points of the output function we couple the probability distributions $\varphi(y)$ across the different query locations $y \in \curlyY$. We first consider a proposal score function $g: \curlyY \to \R^{n\times d_s}$. If we were to compose this function with a map into $d_s$ copies of the probability simplex $\Delta^n$, such as the softmax function $\sigma: \R^n \to \Delta^n$ applied to the rows of $g(y)$, we would obtain the probability distributions
\begin{align*}
    \varphi(y) = \sigma (g(y)).
\end{align*}

The disadvantage of this formulation is that it solely relies on the form of the function $g$ to capture relations between the distributions $\varphi(y)$ across different $y \in \curlyY$. Instead, we introduce the Kernel-Coupled Attention (KCA) mechanism to model these relations by integrating the function $g$ against a coupling kernel $\kappa: \curlyY \times \curlyY \to \R$. This results in the score function,
\begin{align} \label{eq: smoothed g}
    \tilde g(y) = \int_\curlyY \kappa(y,y') g(y')\;dy',
\end{align}
which can be normalized across its rows to form the probability distributions
\begin{align} \label{eq: smoothed phi}
    \varphi(y) = \sigma\left(\int_\curlyY \kappa(y,y') g(y')\;dy'\right).
\end{align}

The form of the kernel $\kappa$ will determine how these distributions are coupled across $y \in \curlyY$. For example, given a fixed $y$, the locations $y'$ where $k(y,y')$ is large will enforce similarity between the corresponding score functions $\tilde g(y)$ and $\tilde g(y')$. If $k$ is a local kernel with a small bandwidth then points $y$ and $y'$ will only be forced to have similar score functions if they are very close together.

\begin{figure}[tp]
\centering
\includegraphics[width=0.65\textwidth]{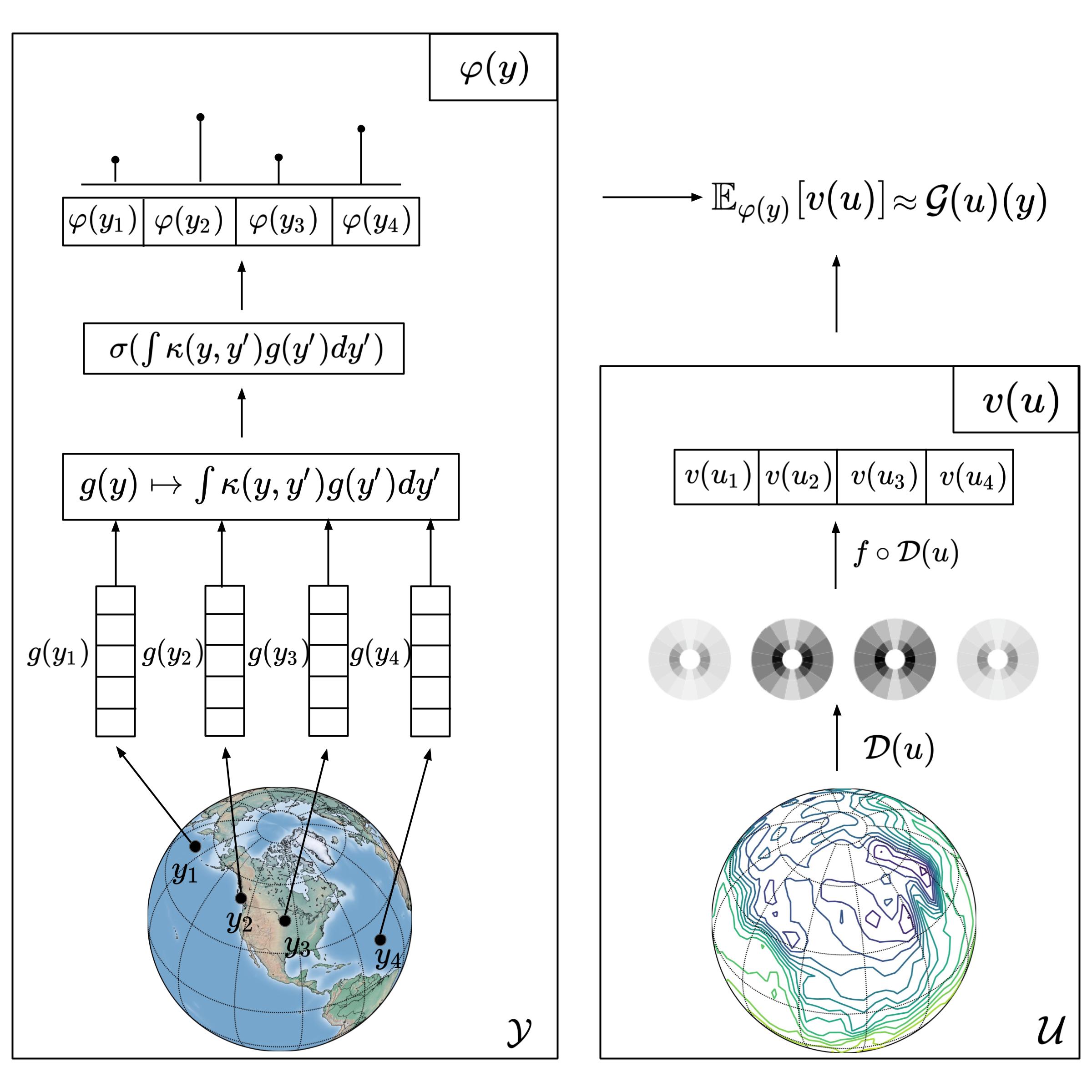}
\caption{Schematic illustration of LOCA: The LOCA method builds a feature representation, $v(u)$, of the input function and averages it with respect to $\varphi(y)$. The transform $\curlyD$ is first applied to the input function to produce a list of features, illustrated by disks in this case, and then a fully-connected network is applied to construct $v(u)$. The score function $g$ is applied to the output query locations $y_i$ together with the softmax function to produce the score vector $\varphi_i$. The $v_i$ and $\varphi_i$ vectors are combined to evaluate the solution at each query location by computing $\E_{\varphi(y)}[v(u)]$ at the last step.}
\label{fig:master_figure}
\end{figure}

\subsection{Formulation of the Coupling Kernel} \label{sec: kernel formulation}

In this section we construct the coupling kernel $\kappa$ that will be used to relate the query distributions as in \eqref{eq: smoothed phi}. We first lift the points $y \in \curlyY$ via a nonlinear parameterized mapping $q_\theta: \curlyY \to \R^l$. We then apply a universal kernel $k: \R^l \times \R^l \to \R$ \cite{micchelli2004kernels} over the lifted space, such as the Gaussian RBF kernel,
\begin{align} \label{eq: RBF}
k(z,z') = \gamma \exp(- \beta \|z-z'\|^2), \quad \gamma, \beta > 0.
\end{align}
Finally, we apply a normalization to the output of this kernel on the lifted points to create a similarity measure. The effect of the normalization is to maintain the relative scale of the proposal score function $g$. Overall, our kernel is defined as
\begin{align} \label{eq: kernel def}
    \kappa(y,y') := \frac{k(q_{\theta}(y),q_{\theta}(y'))}{\left(\int_\curlyY k(q_{\theta}(y),q_{\theta}(z)) dz\right)^{1/2}\left(\int_\curlyY k(q_{\theta}(y'),q_{\theta}(z)) dz\right)^{1/2}}.
\end{align}

By tuning the parameters $\theta$, $\beta$ and $\gamma$ in the functions $q_\theta$ and $k$, the kernel $\kappa$ is able to learn the appropriate measures of similarity between the points in the output function domain $\curlyY$.

\subsection{Input Function Feature Encoding}
\label{sec: Input Function Feature Encoding}

The last architecture choice to be made concerns the functional form of the feature embedding $v(u)$. Here, we construct the map $v$ as a composition of two mappings. The first is a function
\begin{align}\label{eq: input function transform}
    \curlyD: C(\mathcal X,\R^{d_u}) \to \R^d,
\end{align}
that maps an input function $u$ to a finite-dimensional vector $\curlyD(u) \in \R^d$. After creating the $d$-dimensional representation of the input function $\curlyD(u)$, we pass this vector through a function $f$ from a class of universal function approximators, such as fully connected neural networks. The composition of these two operations forms our feature representation of the input function,
\begin{align} \label{eq: input feature map}
    v(u) = f \circ \curlyD (u).
\end{align}

One example for the operator $\curlyD$ is the image of the input function under $d$ linear functionals on $C(\mathcal X; \R^{d_u})$. For example, $\curlyD$ could return the point-wise evaluation of the input function at $d$ fixed points. This would correspond to the action of $d$ translated $\delta$-functionals. The drawback of such an approach is that the model would not be able to accept measurements of the input function at any other locations. As a consequence the input resolution could never vary across forward passes of the model.

Alternatively, if we consider an orthonormal basis for $L^2(\mathcal X; \R^{d_u})$, we could also have $\curlyD$ be the projection onto the first $d$ basis vectors. For example, if we use the basis of trigonometric polynomials the Fast Fourier Transform (FFT) \cite{cooley1965algorithm} allows for efficient computation of these values and can be performed across varying grid resolutions. We could also consider the projection onto an orthogonal wavelet basis \cite{daubechies1988orthogonal}. In the case of complex valued coefficients for these basis functions, the range space dimension of $\curlyD$ would be doubled to account for the real and imaginary part of these measurements.

\subsection{Model Summary} Overall, the forward pass of the proposed model is written as follows, see Figure \ref{fig:master_figure} for a visual representation.
\begin{align} \label{eq: model summary}
\curlyF(u)(y) =  \mathbb{E}_{\varphi(y)}[v(u)] = \sum_{i=1}^n \sigma\left(\int_\curlyY \kappa(y,y') g(y')\;dy'\right)_i \odot v_i(u),
\end{align}
where $\kappa:\mathcal{Y}\times\mathcal{Y}\rightarrow\mathbb{R}$ is the kernel of equation \eqref{eq: smoothed g}, $\sigma$ is the softmax function, $v$ the input feature encoder and $g$ is the proposed score function. Practical aspects related to the parametrization of $\kappa$, $v$ and $g$, as well as the model evaluation and training will be discussed in section \ref{sec:implementation of method}.

In the next sections, we will perform analysis on this model. We will show that under certain architecture choices other models in the literature can be recovered and theoretical guarantees of universal approximation can be proven.

\section{Theoretical Guarantees of Universality} \label{sec: univ}

In this section we give conditions under which the LOCA model is universal. There exist multiple definitions of universality present in the literature, for example see Sriperumbudur \textit{et. al.} \cite{sriperumbudur11a}. To be clear, we formally state the definition we use below.

\begin{definition}[Universality]
Given compact sets $\curlyX \subset \R^{d_x}$, $\curlyY \subset \R^{d_y}$ and a compact set $\curlyU \subset C(\curlyX, \R^{d_u})$ we say a class of operators $\curlyA \ni \curlyF: C(\curlyX, \R^{d_u}) \to C(\curlyY, \R^{d_s})$ is \emph{universal} if it is dense in the space of operators equipped with the supremum norm. In other words, for any continuous operator $\curlyG: C(\curlyX, \R^{d_u}) \to C(\curlyY, \R^{d_s})$ and any $\epsilon > 0$, there exists $\curlyF \in \curlyA$ such that
\begin{align*}
    \sup_{u \in \curlyU} \sup_{y \in \curlyY} \|\curlyG(u)(y) - \curlyF(u)(y)\|_{\R^{d_s}}^2 < \epsilon.
\end{align*}
\end{definition}

To explore the universality properties of our model we note that if we remove the softmax normalization and the kernel coupling, the evaluation of the model can be written as
\begin{align*} 
\curlyF(u)(y) = \sum_{i=1}^n g_i(y) \odot v_i(u).
\end{align*}
The universality of this class of models has been proven in Chen \textit{et. al.} \cite{chen1995universal} (when $d_s = 1$) and extended to deep architectures in Lu \textit{et. al.} \cite{lu2019deeponet}. We will show that our model with the softmax normalization and kernel coupling is universal by adding these components back one at a time. First, the following theorem shows that the normalization constraint $\varphi(y) \in \prod_{k=1}^{d_s} \Delta^n$ does not reduce the approximation power of this class of operators. 

\begin{theorem}[Normalization Preserves Universality] \label{thm: universality}
If $\curlyU \subset C(\curlyX, \R^{d_u})$ is a compact set of functions and $\curlyG: \curlyU \to C(\curlyY, \R^{d_s})$ is a continuous operator with $\curlyX$ and $\curlyY$ compact, then for every $\epsilon > 0$ there exists $n \in \N$, functionals $v_{j,k}: \curlyU \to \R$, for $j \in [n]$, $k \in [d_s]$, and functions $\varphi_j: \curlyY \to \R^{d_s}$ with $\varphi_j(y) \in [0,1]^{d_s}$ and $\sum_{j=1}^n \varphi_j(y) = 1_{d_s}$ for all $y \in \curlyY$ such that
\begin{align*} 
\sup_{u \in \curlyU} \sup_{y \in \curlyY} \left\| \curlyG(u)(y) - \E_{\varphi(y)}[v(u)]\right\|_{\R^{d_s}}^2 < \epsilon.
\end{align*}
\end{theorem}
\begin{proof}
The proof is given in Appendix \ref{sec: main universality proof}.
\end{proof}

It remains to show that the addition of the kernel coupling step for the functions $\varphi$ also does not reduce the approximation power of this class of operators. By drawing a connection to the theory of Reproducing Kernel Hilbert Spaces (RKHS), we are able to state the sufficient conditions for this to be the case. The key insight is that, under appropriate conditions on the kernel $\kappa$, the image of the integral operator in \eqref{eq: smoothed g} is dense in an RKHS $\curlyH_{\kappa}$ which itself is dense in $C(\curlyY, \R^n)$. This allows \eqref{eq: smoothed phi} to approximate any continuous function $\varphi: \curlyY \to \prod_{i=1}^{d_s} \Delta^n$ and thus maintains the universality guarantee of Theorem \ref{thm: universality}.

\begin{proposition}[Kernel Coupling Preserves Universality] \label{prop: phi univ}
Let $\kappa: \curlyY \times \curlyY \to \R$ be a positive definite and symmetric universal kernel with associated RKHS $\curlyH_\kappa$ and define the integral operator
\begin{align*}
    T_\kappa: &C(\curlyY, \R^n) \to C(\curlyY, \R^n), \\
                &f \mapsto \int_\curlyY \kappa(y,z) f(z) dz.
\end{align*}
If $\curlyA \subseteq C(\curlyY, \R^n)$ is dense, then $T_\kappa(\curlyA) \subset C(\curlyY, \R^n)$ is also dense.
\end{proposition}

The statement of Proposition \ref{prop: phi univ} requires that the kernel $\kappa$ be symmetric, positive definite, and universal.  We next show that by construction it will always be symmetric and positive definite, and under an assumption on the feature map $q$ it will additionally be universal.

\begin{proposition}[Universality of the Kernel $\kappa$]\label{prop: PD symm}
    The kernel defined in \eqref{eq: kernel def} is positive definite and symmetric. Further, if $q$ is injective, it defines a universal RKHS.
\end{proposition}
\begin{proof}
The proof is provided in Appendix \ref{sec: symmetry PD proof}.
\end{proof}

Lastly, we present a result showing that a particular architecture choice for the input feature encoder $v$ also preserves universality.  We show that if there is uniform convergence of spectral representations of the input, projections onto these representations can be used to construct a universal class of functionals on $C(\curlyX, \R^{d_u})$.
\begin{proposition}[Spectral Encoding Preserves Universality] \label{prop: v univ}
    Let $\curlyA_d \subset C(\R^d, \R^n)$ be a set of functions dense in $C(\R^d, \R^n)$, and $\{e_i\}_{i=1}^\infty$ a set of basis functions such that for some compact set $\curlyU \subseteq C(\curlyX, \R^{d_u})$, $\sum_{i=1}^\infty \langle u, e_i \rangle_{L^2} e_i$ converges to $u$ uniformly over $\curlyU$. Let $\curlyD_d: \curlyU \to \R^d$ denote the projection onto $\{e_1, \ldots, e_d\}$. Then for any continuous functional $h: \curlyU \to \R^n$, and any $\epsilon > 0$, there exists $d$ and $f \in \curlyA_N$ such that
    $$\sup_{u \in \curlyU} \big\| h(u) - f \circ \curlyD_d(u)\big\| < \epsilon.$$
\end{proposition}
\begin{proof}
The proof is provided in Appendix \ref{sec: v unif proof}.
\end{proof}
For example, if our compact space of input functions $\curlyU$ is contained in $C^1(\curlyX, \R^{d_u})$, and $\curlyD$ is a projection onto a finite number of Fourier modes, the  architecture proposed in equation \eqref{eq: input feature map} is expressive enough to approximate any functional from $\curlyU \to \R$, including those produced by the universality result stated in Theorem \ref{thm: universality}.

\section{Implementation Aspects}
\label{sec:implementation of method}

To implement our method, it remains to make a choice of discretization for computing the integrals required for updating the KCA weights $\varphi(y)$, as well as a choice for the input function feature encoding $v(u)$. Here we address these architecture choices, and provide an overview of the proposed model's forward evaluation. 

\subsection{Computation of the Kernel Integrals}\label{sec:computation of score function}

To compute the kernel-coupled attention weights $\varphi(y)$, we are required to evaluate integrals over the domain $\curlyY$ in \eqref{eq: smoothed g} and \eqref{eq: kernel def}. Adopting an unbiased Monte-Carlo estimator using $P$ points $y_1, \ldots, y_P \in \curlyY$, we can use the approximations
\begin{align*}
    \int_\curlyY \kappa(y,y') g(y') \approx \frac{\mathrm{vol}(\curlyY)}{P}\sum_{i=1}^P \kappa(y,y_i) g(y_i),
\end{align*}
for equation \eqref{eq: smoothed g}, and
\begin{align*}
    \int_\curlyY k(q(y),q(z)) dz \approx \frac{\mathrm{vol}(\curlyY)}{P}\sum_{i=1}^P k(q(y),q(y_i)),
\end{align*}
for use in equation \eqref{eq: kernel def}. Note that due to the normalization in $\kappa$, the $\mathrm{vol}(\curlyY)$ term cancels out. In practice, we allow the query point $y$ to be one of the points $y_1, \ldots, y_P$ used for the Monte-Carlo approximation.

When the domain $\curlyY$ is low dimensional, as in many physical problems, a Gauss-Legendre quadrature rule with weights $w_i$ can provide an accurate and efficient alternative to Monte Carlo approximation. Using $Q$ Gauss-Legendre nodes and weights, we can approximate the required integrals as
\begin{align*}
    \int_\curlyY \kappa(y,y') g(y') dy' \approx \sum_{i=1}^Q w_i \kappa(y,y'_i) g(y'_i),
\end{align*}
for equation \eqref{eq: smoothed g} and 
\begin{align*}
    \int_\curlyY k(q(y),q(z))  dz \approx \sum_{i=1}^Q w_i k(q(y),q(z_i)),
\end{align*}
for use in equation \eqref{eq: kernel def}. 

If we restrict the kernel $\kappa$ to be translation invariant, there is another option for computing these integrals. As in Li \textit{et. al.} \cite{li2020fourier}, we could take the Fourier transform of both $\kappa$ and $g$, perform a point-wise multiplication in the frequency domain, followed by an inverse Fourier transform. However, while in theory the discrete Fourier transformation could be performed on arbitrarily spaced grids, the most available and computationally efficient implementations rely on equally spaced grids. We prefer to retain the flexibility of arbitrary sets of query points $y$ and will therefore not pursue this alternate approach. In Section \ref{sec:Results}, we will switch between the Monte-Carlo and quadrature strategies depending on the problem at hand. 

\subsection{Positional Encoding of Output Query Locations}
\label{sec:APPositionalEncoding}

We additionally adopt the use of positional encodings, as they have been shown to improve the performance of attention mechanisms. For encoding the output query locations, we are motivated by the positional encoding in Vaswani \textit{et. al.} \cite{vaswani2017attention}, the harmonic feature expansion in Di \textit{et. al.} \cite{di2021deeponet}, and the work of Wang \textit{et. al.} \cite{wang2019translating} for implementing the encoding to more than one dimensions. The positional encoding for a one dimensional query space is given by
\begin{equation} \label{eq: positional encoding}
    \begin{split}
        &e(y^i,2j+(i-1)H) = \cos ( 2^j \pi y^i ) \\
        &e(y^i,2j+1+(i-1)H) = \sin ( 2^j \pi y^i ),
    \end{split}
\end{equation}
where $H$ the number of encoding coefficients, $j=1,...,H/2$, $y^i$ the query coordinates in different spatial dimensions and  $i=1,...,d_y$. In contrast to Vaswani \textit{et. al.} \cite{vaswani2017attention} we consider the physical position of the elements of the set $y$ as the position to encode instead of their index position in a given list, as the index position in general does not have a physically meaningful interpretation. 

\subsection{Wavelet Scattering Networks as a Spectral Encoder}
While projections onto an orthogonal basis allows us to derive a universality guarantee for the architecture, there can be some computational drawbacks. For example, it is known that the Fourier transform is not always robust to small deformations of the input \cite{mallat2012group}. More worrisome is the lack of robustness to noise corrupting the input function. In real world applications it will often be the case that our inputs are noisy, hence, in practice we are motivated to find an operator $\curlyD$ with stronger continuity with respect to these small perturbations.

To address the aforementioned issues, we make use of the scattering transform \cite{bruna2013invariant}, as an alternate form for the operator $\curlyD$. The scattering transform maps an input function to a sequence of values by alternating wavelet convolutions and complex modulus operations \cite{bruna2013invariant}. To be precise, given a mother wavelet $\psi$ and a finite discrete rotation group $G$, we denote the wavelet filter with parameter $\lambda = (r, j) \in G \times \mathbb{Z}$ as
\begin{align*}
\psi_{\lambda}(u) = 2^{d_xj}\psi(2^j r^{-1} x).
\end{align*}
Given a path of parameters $p= (\lambda_1, \ldots, \lambda_m)$, the scattering transform is defined by the operator 
\begin{align} \label{eq: path scattering}
    S[p]u =  ||||u \star \psi_{\lambda_1}| \star \psi_{\lambda_2} |\cdots | \star \psi_{\lambda_m}| \star \phi(x),
\end{align} 
where $\phi(x)$ is a low pass filter. We allow the empty path $\emptyset$ as a valid argument of $S$ with $S[\emptyset]u = u \star \phi$. As shown in Bruna \textit{et. al.} \cite{bruna2013invariant}, this transform is Lipschitz continuous with respect to small deformations, while the modulus of the Fourier transform is not. This transform can be interpreted as a deep convolutional network with fixed filters and has been successfully applied in multiple machine learning contexts \cite{oyallon2017scaling, chang2015learning}. Computationally, the transform returns functions of the form \eqref{eq: path scattering} sampled at points in their domain, which we denote by $\hat S[p](u)$.

By choosing $d$ paths $p_1, \ldots, p_d$, we may define the operator $\curlyD$ as
\begin{align*}
    \curlyD(u) = \left(\hat S[p_1](u), \ldots, \hat S[p_d](u)\right)^{\top}.
\end{align*}

In practice, the number of paths used is determined by three parameters: $J$, the maximum scale over which we take a wavelet transform; $L$, the number of elements of the finite rotation group $G$, and $m_0$, the maximum length of the paths $p$. 
While Proposition \ref{prop: v univ} does not necessarily apply to this form of $\curlyD$, we find that empirically this input encoding gives the best performance. 

\subsection{Loss Function and Training} The proposed model is trained by minimizing the empirical risk loss over the available training data pairs,
\begin{align}\label{eq:LOCA loss}
\curlyL(\theta) = \frac{1}{N} \sum_{i=1}^N \sum_{\ell=1}^P ( s^i(y^i_\ell) -  \curlyF_{\theta}(u^i)(y^i_\ell) )^2,
\end{align}
where $\theta = (\theta_q, \theta_f ,\theta_g )$ denotes all trainable model parameters.
This is the simplest choice that can be made for training the model. Other choices may include weighting the mean square error loss using the $\mathcal{L}_1$ norm of the ground truth output \cite{di2021deeponet, wang2021improved}, or employing a relative $\mathcal{L}_2$ error loss \cite{li2020fourier}. The minimization is performed via stochastic gradient descent updates, where the required gradients of the loss with respect to all the trainable model parameters can be conveniently computed via reverse-mode automatic differentiation.

\subsection{Implementation Overview}

Algorithm \ref{alg:implementation_algor} provides an overview of the steps required for implementing the LOCA method. The training data set is first processed by passing the input functions through a wavelet scattering network \cite{bruna2013invariant}, followed by applying a positional encoding to the query locations and the quadrature/Monte-Carlo integration points. The forward pass of the model is evaluated and gradients are computed for use with a stochastic gradient descent optimizer. After training, we make one-shot predictions for super-resolution grids, and we compute the relative $\mathcal{L}_2$ error between the ground truth output and the prediction.

\begin{algorithm}
\caption{Implementation summary of the LOCA method}\label{alg:implementation_algor}
\begin{algorithmic}
\Require 
\LState $\bullet$ Input/output function pairs $\{u^i ,s^i\}_{i=1}^N$.
\LState $\bullet$ Query locations $y^i$ for evaluating $s^i$.
\LState $\bullet$ Quadrature points $z^i$.
\LState {\bf Pre-processing:} 
\LState $\bullet$ Apply transformation \eqref{eq: input feature map} on the input function to get $\hat{u}$, the input features.
\LState $\bullet$ Apply positional encoding \eqref{eq: positional encoding} to query coordinates $y,z$, to get $\hat{y}, \hat{z}$.
\LState $\bullet$ Choose the network architectures for functions $q_{\theta_q}$, $f_{\theta_f}$, and $g_{\theta_g}$. 
\LState $\bullet$ Initialize the trainable parameters $\theta = (\theta_q, \theta_f ,\theta_g )$, and choose a learning rate $\eta$.
\LState {\bf Training:}
\For{$i=0$ to $I$}
\State Randomly select a mini-batch of $(\hat{u}, \hat{y}, \hat{z}, s)$.
\State Evaluate $g_{\theta_g}(q_{\theta_q}(\hat{z}))$. 
\State Compute the Coupling Kernel $\kappa(q_{\theta_q}(\hat{y}), q_{\theta_q}(\hat{z}))$ \eqref{eq: kernel def}.
\State Numerically approximate the KCA \eqref{eq: smoothed g} and compute $\varphi(y)$.
\State Evaluate $f_{\theta_f}(\hat{u})$, as in Equation \eqref{eq: input feature map}.
\State Evaluate the expectation \eqref{eq: model summary} and get $s^*$, the model prediction.
\State Evaluate the training loss \eqref{eq:LOCA loss} and compute its gradients $\nabla_{\theta}\curlyL(\theta_i)$.
\State Update the trainable parameters via stochastic gradient descent: $\theta_{i+1} \leftarrow \theta_{i} - \eta \nabla_{\theta}\curlyL(\theta_i)$.
\EndFor
\end{algorithmic}
\end{algorithm}

\section{Connections to Existing Operator Learning Methods} \label{sec: related methods}

In this section, we provide some insight on the connections between our method and similar operator learning methods.

\subsection{DeepONets}
Note that if we identify our input feature map, $v(u)$, with the DeepONet's branch network, and the location dependent probability distribution, $\varphi(y)$, with the DeepONet's trunk network, then the last step of both models is computed the same way. We can recover the DeepONet architecture from our model under three changes to the architecture in the forward evaluation. First, we would remove the normalization step in the construction of $\varphi$. Next, we remove the KCA mechanism that is applied to the candidate score function $g$ (equivalently we may fix the kernel $\kappa$ to be $\delta$-distributions along the diagonal). Finally, in the construction of the input feature map $v(u)$, instead of the scattering transform we would act on the input with a collection of $\delta$ distributions at the fixed sensor locations.  These differences between DeepONets and LOCA result in increased performance of our model, as we will see in Section \ref{sec:Results}.

\subsection{Neural Operators}
The connection between Neural Operators and DeepONets has been presented in Kovachki \textit{et. al.} \cite{kovachkineural}, where it is shown that a particular choice of neural operator architecture produces a DeepONet with an arbitrary trunk network and a branch network of a certain form. In particular, a Neural Operator layer has the form,
\begin{align} \label{eq: neural op}
    v^{(\ell+1)}(z) = \sigma \left(W^{(
    \ell)}v^{(\ell)}(z) + \int_{\mathcal Z} k^{(\ell)}(s, z)v^{(\ell)}(s)ds\right),
\end{align}
where here $\sigma$ is a point-wise nonlinearity. It is shown in Kovachki \textit{et. al.} \cite{kovachkineural} that this architecture can be made to resemble a DeepONet under the following choices. First, set $W^{(\ell)} = 0$. Next, lift the input data to $n$ tiled copies of itself and choose a kernel $k$ that is separable in $s$ and $z$. If the output of the layer is then projected back to the original dimension by summing the coordinates, the architecture resembles a DeepONet.

The correspondence between our model and DeepONets described above allows us to transitively connect our model to Neural Operators as well. We additionally note that the scattering transform component of our architecture can be viewed as a collection of multiple-layer Neural Operators with fixed weights. Returning to \eqref{eq: neural op}, when $W^{(\ell)} = 0$ for all $\ell$, the forward pass of the architecture is a sequence of integral transforms interleaved with point-wise nonlinearities. Setting $\sigma$ to be the complex modulus function and $k^{(\ell)}$ to be a wavelet filter $\psi_{\lambda_\ell}$ we may write
\begin{align*}
    v^{(\ell+1)} = |v^{\ell} \ast \psi_\lambda|.
\end{align*}
When we compose $L$ of these layers together, we recover \eqref{eq: path scattering} up to the application of the final low pass filter (again a linear convolution)
\begin{align*}
    v^{(L)} = ||||u \ast \psi_{\lambda_1}| \ast \psi_{\lambda_2} |\cdots | \ast \psi_{\lambda_L}|.
\end{align*}

Thus, we may interpret the scattering transform as samples from a collection of Neural Operators with fixed weights. This connection between the scattering transform and convolutional neural architectures with fixed weights was noticed during the original formulation of the wavelet scattering transform by Bruna \textit{et. al.} \cite{bruna2013invariant}, and thus also extends to Neural Operators via the correspondence between Neural Operators and (finite-dimensional) convolutional neural networks \cite{kovachkineural}.

A key difference between our model and Neural Operators is how the kernel integral transform is applied. In the Neural Operator, it is applied directly to the input and the output functions of the internal layers, while in LOCA the kernel acts only on a score function of the output domain $\curlyY$, as in \eqref{eq: smoothed g}.

\subsection{Other Attention-Based Architectures}
Here we compare our method with two other recently proposed attention-based operator learning architectures. The first is the Galerkin/Fourier Transformer \cite{cao2021choose}. This method operates on a fixed input and output grid, and most similarly represents the original sequence-to-sequence Transformer architecture \cite{vaswani2017attention} with different choices of normalization. As in the original sequence-to-sequence architecture, the attention weights are applied across the indices (sensor locations) of the input sequence. By contrast, in our model the attention mechanism is applied to a finite-dimensional feature representation of the input that is not indexed by the input function domain. Additionally, our attention weights are themselves coupled over the domain $\curlyY$ via the KCA mechanism \eqref{eq: smoothed phi} as opposed to being defined over the input function domain in an uncoupled manner.

A continuous attention mechanism for operator learning was also proposed as a special case of Neural Operators in Kovachki \textit{et. al.} \cite{kovachkineural}. There, it was noted that if the kernel in the Neural Operator was (up to a linear transformation) of the form
\begin{align*}
    k(v(x),v(y)) = \left( \int \exp\left(\frac{\langle Av(s), Bv(y)\rangle}{\sqrt{m}}\right)ds\right)^{-1}\exp\left(\frac{\langle Av(x), Bv(y)\rangle}{\sqrt{m}}\right),
\end{align*}
with $A,B \in \R^{m \times n}$, then the corresponding Neural Operator layer can be interpreted as the continuous generalization of a transformer block. Further, upon discretization of the integral this recovers exactly the sequence-to-sequence discrete Transformer model.

The main difference of this kind of continuous transformer with our approach is again how the attention mechanism is applied to the inputs. The Neural Operator Transformer is similar to the Galerkin/Fourier Transformer in the sense that the attention mechanism is applied over the points of the input function itself, whereas our model first creates a different finite dimensional feature representation of the input function which the attention is applied to. We note that our model does make use of attention weights defined over a continuous domain, but it is the domain of the output functions $\curlyY$ as opposed to $\curlyX$. The coupling of the attention weights as a function of the output query in \eqref{eq: smoothed phi} with the kernel in \eqref{eq: kernel def} can be interpreted as a kind of un-normalized continuous self-attention mechanism where we view the query space $\curlyY$ as its own input space to generate the attention weights $\varphi(y)$.

\section{Experimental Results}\label{sec:Results}

\begin{figure} 
\centering
\includegraphics[width=\textwidth]{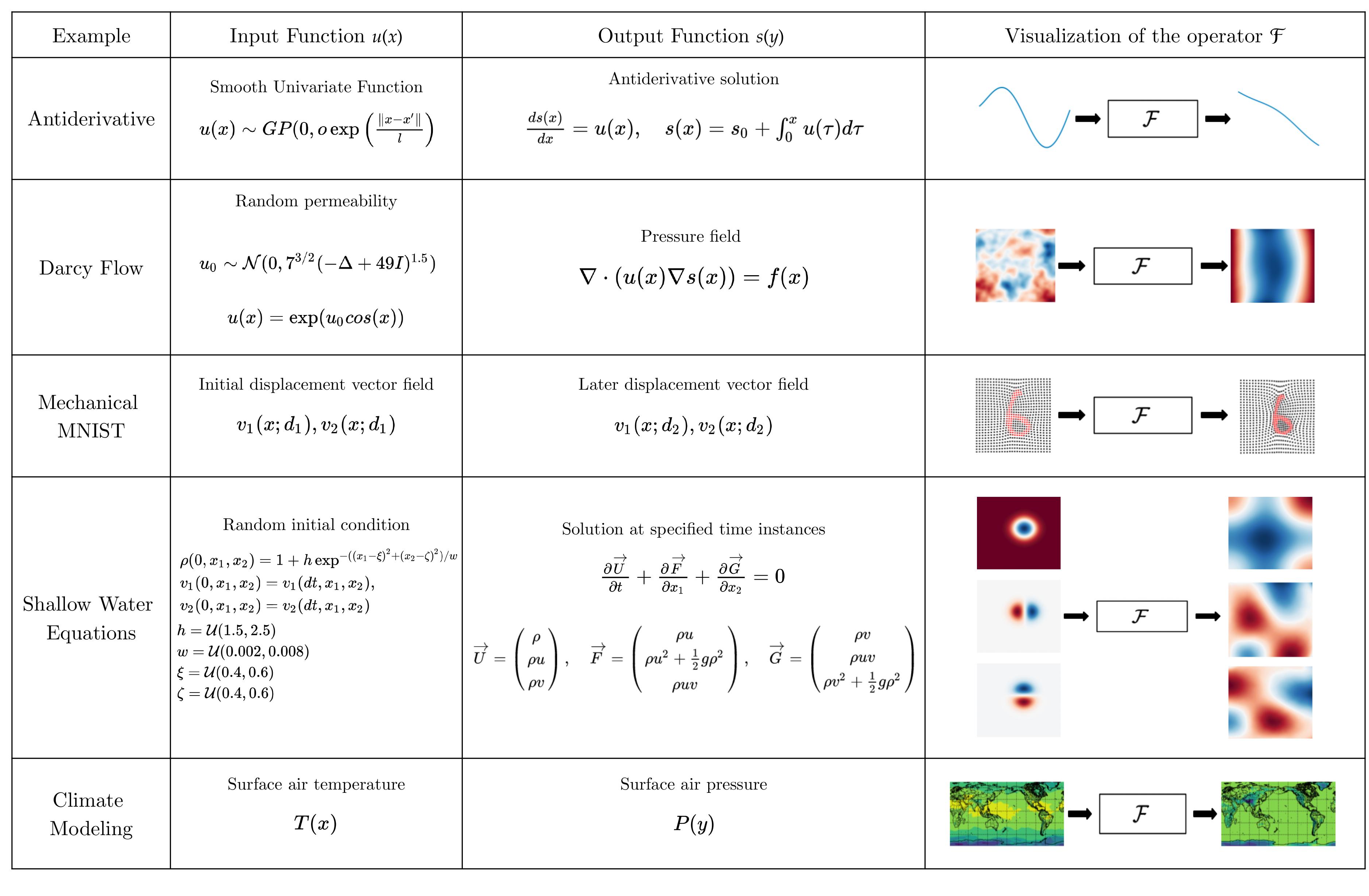}
\caption{ A schematic visualization of the operator learning benchmarks considered in this work. Shown are the input/output function and a description of their physical meaning, as well as the operator that we learn for each example. In the Mechanical MNIST example, for visual clarity we do not present the map that the model is actually learning, which is the displacement in the vertical and the horizontal directions, but the position of each pixel under a specified displacement. See Appendix Section \ref{sec:APexperiments} for more details on the data set generation.}
\label{fig:example_table}
\end{figure}

In this section we provide a comprehensive collection of experimental comparisons designed to assess the performance of the proposed LOCA model against two state of the art operator learning methods, the Fourier Neural Operator (FNO) \cite{li2020fourier} and the DeepONet (DON) \cite{lu2019deeponet}. We will show that our method requires less labeled data than competing methods, is robust against noisy data and randomness in the model initialization, has a smaller spread of errors over testing data sets, and is able to successfully generalize in out-of-distribution testing scenarios. Evidence is provided for the following numerical experiments, see Figure \ref{fig:example_table} for a visual description.

\begin{itemize}
    \item \textbf{Antiderivative:} Learning the antiderivative operator given multi-scale source terms.
    \item \textbf{Darcy Flow:} Learning the solution operator of the Darcy partial differential equation, which models the pressure of a fluid flowing through a porous medium with random permeability.
    \item \textbf{Mechanical MNIST:} Learning the mapping between the initial and final displacement of heterogeneous block materials undergoing equibiaxial extension.
    \item \textbf{Shallow Water Equations:} Learning the solution operator for a partial differential equation describing the flow below a pressure surface in a fluid with reflecting boundary conditions.
    \item \textbf{Climate modeling:} Learning the mapping from the air temperature field over the Earth's surface to the surface air pressure field, given sparse measurements.
\end{itemize}

For all experiments the training data sets will take the following form. For each of the $N$ input/output function pairs, $(u^i, s^i)$, we will consider $m$ discrete measurements of each input function at fixed locations, $(u^i(x^i_1), \ldots, u^i(x^i_m))$, and $M$ available discrete measurements of each output function $(s^i(y^i_1), \ldots, s^i(y^i_M))$, with the query locations $\{y^i_{\ell}\}_{\ell=1}^M$ potentially varying over the data set. Out of the $M$ available measurement points $\{y^i_{\ell}\}_{\ell=1}^M$ for each output function $s^i$, we consider the effect of taking only $P$ of these points for each input/output pair. For example, if we use $10\%$ of labeled data, we set $P = \lfloor M/10 \rfloor$ and build a training data set where each example is of the form $(\{u^i(x^i_j)\}_{j=1}^m, \{s^i(y_\ell)\}_{\ell=1}^{P})$. We round the percentages to the nearest integer or half-integer for clarity. We present details on the input and output data construction, as well as on the different problem formulations in Section \ref{sec:APexperiments} of the Appendix.

In each scenario the errors are computed between both the models output and ground truth at full resolution. Throughout all benchmarks, we employ Gaussian Error Linear unit activation functions (GELU) \cite{hendrycks2016gaussian}, and initialize all networks using the Glorot normal scheme \cite{glorot2010understanding}. All networks are trained via  mini-batch stochastic gradient descent using the Adam optimizer with default settings \cite{kingma2014adam}. The detailed hyper-parameter settings, the associated number of parameters for all examples, the computational cost, and other training details are provided in Appendix \ref{sec:training_details}. All code and data accompanying this manuscript will be made publicly available at \url{https://github.com/PredictiveIntelligenceLab/LOCA}.

\subsection{Data Efficiency}
\label{sec:Data Efficiency}

In this section we investigate the performance of our model when the number of labeled output function points is small. In many applications labeled output function data can be scarce or costly to obtain. Therefore, it is desirable that an operator learning model is able to be successfully trained even without a large number of output function measurements. We investigate this property in the Darcy flow experiment by gradually increasing the percentage of labeled output function measurements used per input function example. Next, we compare the performance of all models for the Shallow Water benchmark in the small data regime. Lastly, we demonstrate that the proposed KCA weights provide additional training stability specifically in the small data regime. One important aspect of learning in the small data regime is the presence of outliers in the error statistics, which quantify the worst-case-scenario predictions. In each benchmark we present the following error statistics across the testing data set: the error spread around the median, and outliers outside the third quantile.

\begin{figure}
\centering
\includegraphics[width=.7\textwidth]{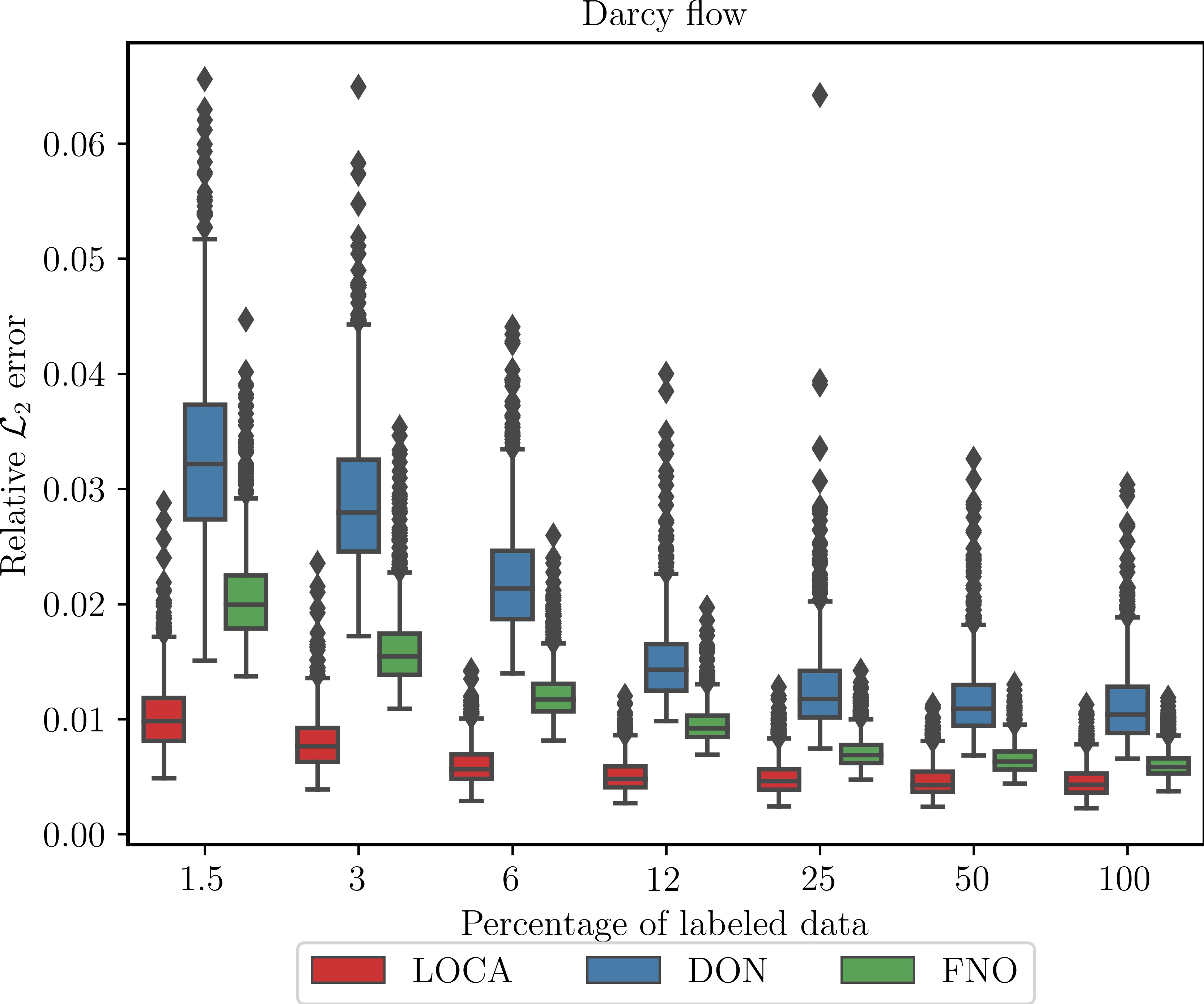}
\caption{ (Data Efficiency) Relative $\mathcal{L}_2$ error boxplots for the solution of Darcy flow: We present the error statistics for the case of the Darcy flow in the form of boxplots for the case where we train on $[1.5,...,100] \%$ of the available output function measurements per example. We observe that our model presents fast convergence to a smaller median error than the other methods and the error spread is more concentrated around the median with fewer outliers.}
\label{fig:darcy_box_plot}
\end{figure}

Figure \ref{fig:darcy_box_plot} shows the effect of varying the percentage of labeled output points used per training example in the Darcy flow prediction example. The box plot shows the distribution of errors over the test data set for each model. We see that the proposed LOCA model is able to achieve low prediction errors even with $1.5\%$ of the available output function measurements per example.  It also has a consistently smaller spread of errors with fewer outliers across the test data set in all scenarios. Moreover, when our model has access to $6 \%$ of the available output function measurements it achieves lower errors against both the DON and FNO trained with any percentage (up to $100 \%$) of the total available labeled data.

\begin{figure}[t]
\centering
\includegraphics[width=\textwidth]{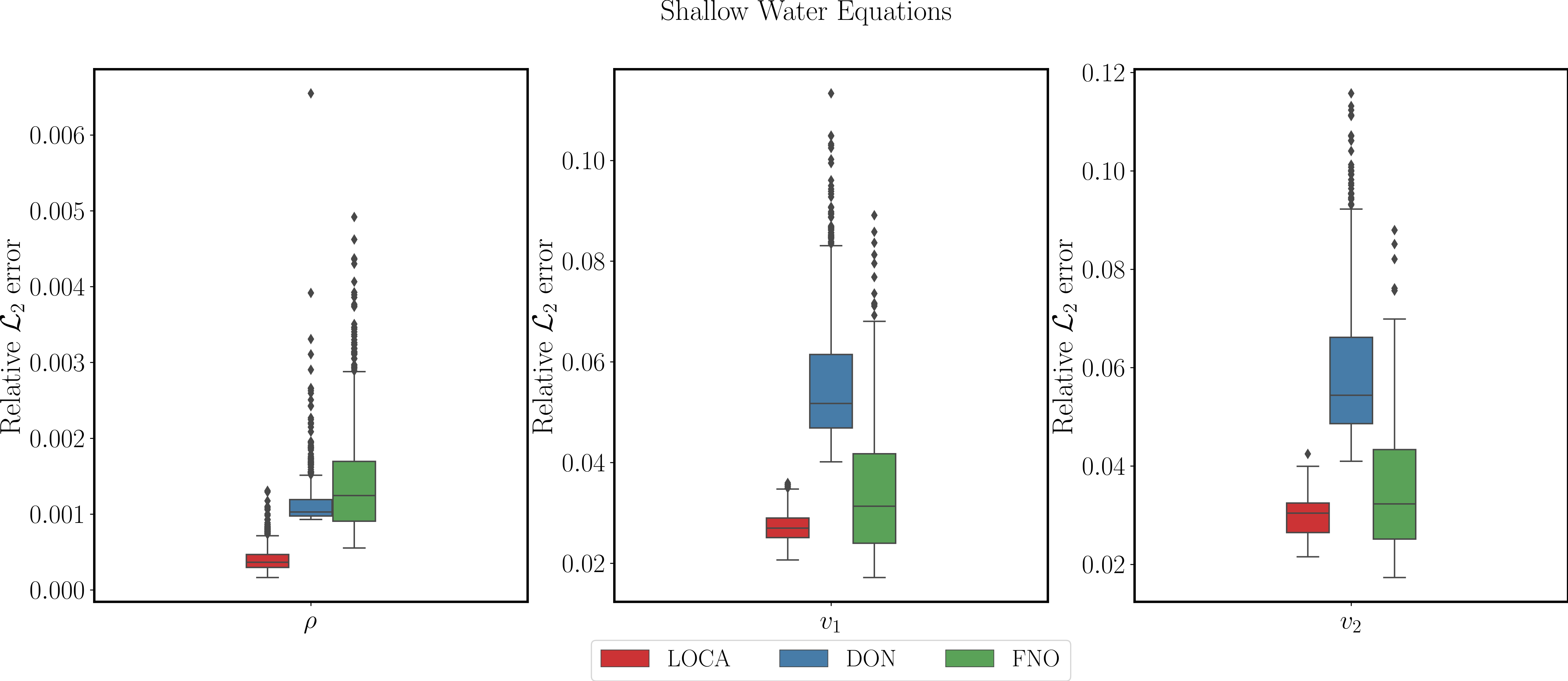}
\caption{ (Data Efficiency)  Relative $\mathcal{L}_2$ error boxplots for the solution of the Shallow Water equations: We present the errors for each different predicted quantity of the Shallow Water equations. On the left, we present the $\rho$ quantity which is the height of the water, and $v_1$ and $v_2$ which are the two components of the fluid velocity vector. We observe that LOCA achieves higher accuracy, and presents fewer outliers and more concentrated error spread compared its competitors.}
\label{fig:sw_box}
\end{figure}

Figure \ref{fig:sw_box} shows the spread of errors across the test data set for the Shallow Water benchmark when the LOCA model is trained on $2.5\%$ of the available labeled data per input-output function pair. We observe that our model outperforms DON and FNO in predicting the wave height, $\rho$, and provides similar errors to the FNO for the two velocity components, $v_1$ and $v_2$. Despite the fact that the two methods perform in a similar manner for the median error, LOCA consistently provides a much smaller standard deviation of errors across the test data set, as well as far fewer outliers. 

We hypothesize that the ability of our model to successfully learn from fewer output function measurements stems from the KCA mechanism used in constructing $\varphi(y)$. By coupling the values of the output function in this way, the model is able to learn the global behavior of the output functions with fewer example points. To demonstrate this, we use a low percentage of output function measurements, and train the LOCA model with and without the KCA step. Table \ref{tab:kernelNokernel_box_plot_errors} shows the result for the case where we use the KCA weights, and the case where we do not. We consider $1.5\%$ of the available labeled data and lower the amount of samples from $N=1,000$ to $N=200$. When KCA is removed, the training becomes unstable and results in a high testing error. With the KCA step for $\varphi$ included, we see that the model still performs well in this small data regime.

\begin{table}
\centering
\resizebox{1\textwidth}{!}{%
\begin{tabular}{|l|l|l|l|l|}
\hline
\backslashbox{Model}{Test Error Metric}&  Mean & Standard deviation & Minimum  & Maximum \\ \hline
LOCA without KCA & 0.463 & 0.184 &  0.406  & 3.023 \\ \hline
LOCA with KCA & \bf{0.017} & \bf{0.004} & \bf{0.008} & \bf{0.041} \\ \hline
\end{tabular}
}
\caption{ (Data Efficiency)  Performance of LOCA with and without KCA: We present the mean, the standard deviation, the minimum and the maximum relative $\curlyL_2$ errors for the Darcy equation with and without KCA when using $1.5\%$ of available output function measurements per training example. We see that the presence of KCA in the model guarantees stability in the training and results in small testing error.}
\label{tab:kernelNokernel_box_plot_errors}
\end{table}

\subsection{Robustness}
\label{sec:Robustness}

Operator learning can be a powerful tool for cases where we have access to clean simulation data for training, but wish to deploy the model on noisy experimental data. Alternatively, we may have access to noisy data for training and want to make predictions on noisy data as well. We will quantify the ability of our model to handle noise in the data by measuring the percentage increase in mean error clean to noisy data scenarios. For all experiments in this section, we consider $7 \%$ of the available labeled data. 

\begin{figure}
\centering
\includegraphics[width=\textwidth]{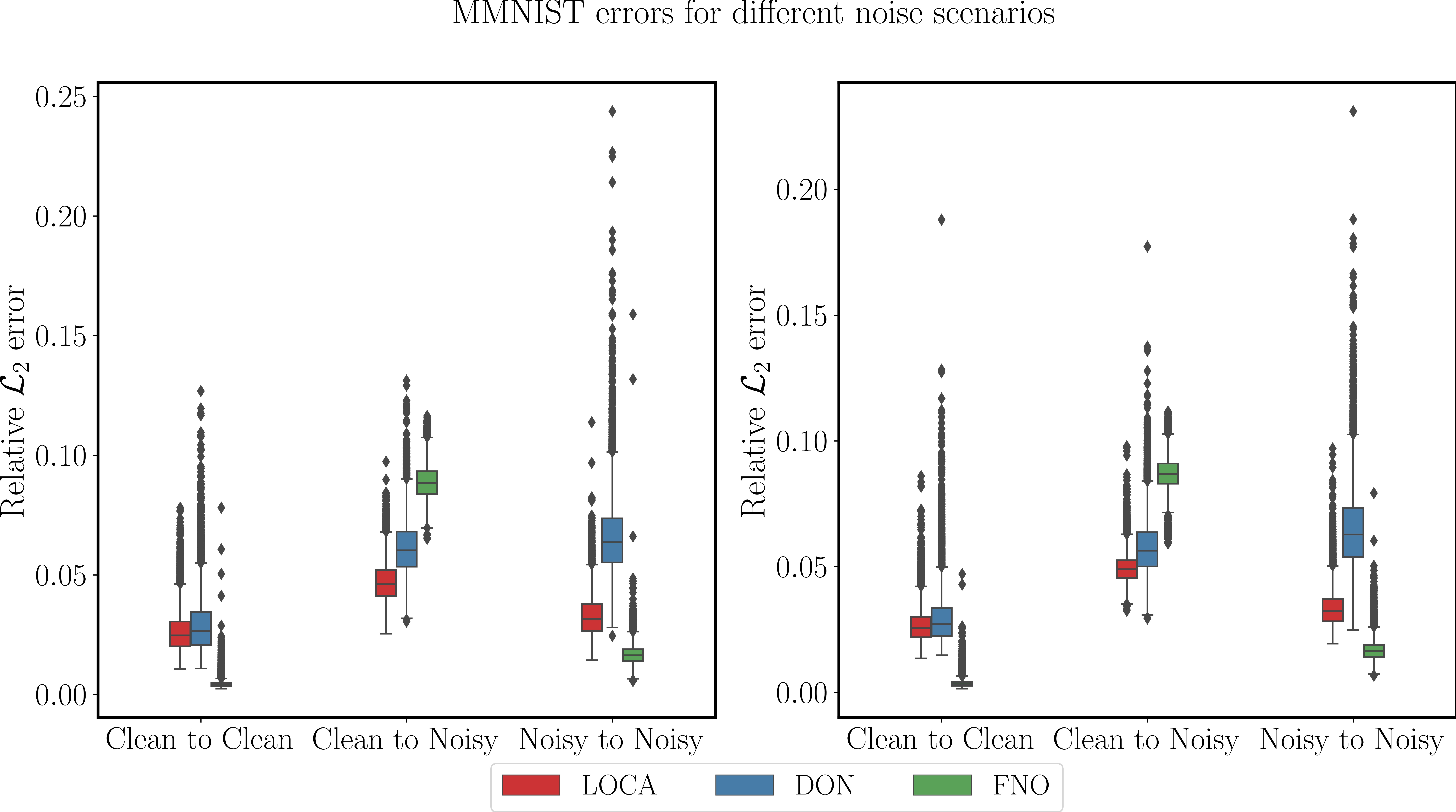}
\caption{(Robustness)  Relative $\mathcal{L}_2$ error boxplots for the Mechanical MNIST benchmark with noisy data:  The left figure gives the distribution of errors for the displacement in the horizontal axis, $v_1$, and the right figure gives the displacement in the vertical axis $v_2$. For all cases we consider $7 \%$ of the whole training data set as labeled data used during training.}
\label{fig:mmnist_box_plot_errors_ccnn}
\end{figure}

We use the Mechanical MNIST benchmark to investigate the robustness of our model with respect to noise in the training and testing data. We consider three scenarios: one where the training and the testing data sets are clean, one where the training data set is clean, but the output data set is corrupted Gaussian noise sampled from $\mathcal{N}(0, .15I)$, and one where both the input and the output data sets are corrupted by Gaussian noise sampled from $\mathcal{N}(0, .15I)$. In Figure \ref{fig:mmnist_box_plot_errors_ccnn} we present the distribution of errors across the test data set for each noise scenario. We observe that for the case where both the training and the testing data are clean, the FNO achieves the best performance. In the scenario where the training data set is clean but the testing data set is noisy, we observe a percentage increase to the approximation error of all methods. 

For the Clean to Noisy scenario the approximation error of the FNO method is increased by $1,930\%$  and $2,238 \%$ for the displacement in the horizontal and vertical directions, respectively. For the DON method, the percentage increase is $112 \%$ and $96 \%$ for the displacement in the horizontal and vertical directions (labeled as $v_1$ and $v_2$), respectively. For the LOCA method the percentage increase is $80 \%$ and $85 \%$ for the displacement in the horizontal and vertical directions, respectively. For the Noisy to Noisy scenario the approximation error of the FNO method is increased by $280 \%$ and $347 \%$ for the displacement in the horizontal and vertical directions, respectively. For the DON method, the percentage increase is $128 \%$ and $120 \%$, and for LOCA is only $26 \%$ and $25 \%$ for each displacement component, respectively. We present the mean prediction error for each scenario and the corresponding percentage error increase in Table \ref{tab:mmnist_percentage_increase}. 

We observe that even though the FNO is very accurate for the case where both training and test data sets are clean, a random perturbation of the test data set can cause a huge decrease in accuracy. On the other hand, even though the DON method presents similar accuracy as our model in the clean to clean case, the standard deviation of the error is greater and its robustness to noise is inferior. LOCA is clearly superior in the case where the testing data are corrupted with Gaussian noise. We again emphasise that the metric in which we assess the performance is not which method has the lowest relative prediction error, but which method presents the smallest percentage increase in the error when noise exists in testing (and training in the case of Noisy to Noisy) data compared to the case where there exist no noise. 

\begin{figure}
\centering
\includegraphics[width=0.9\textwidth]{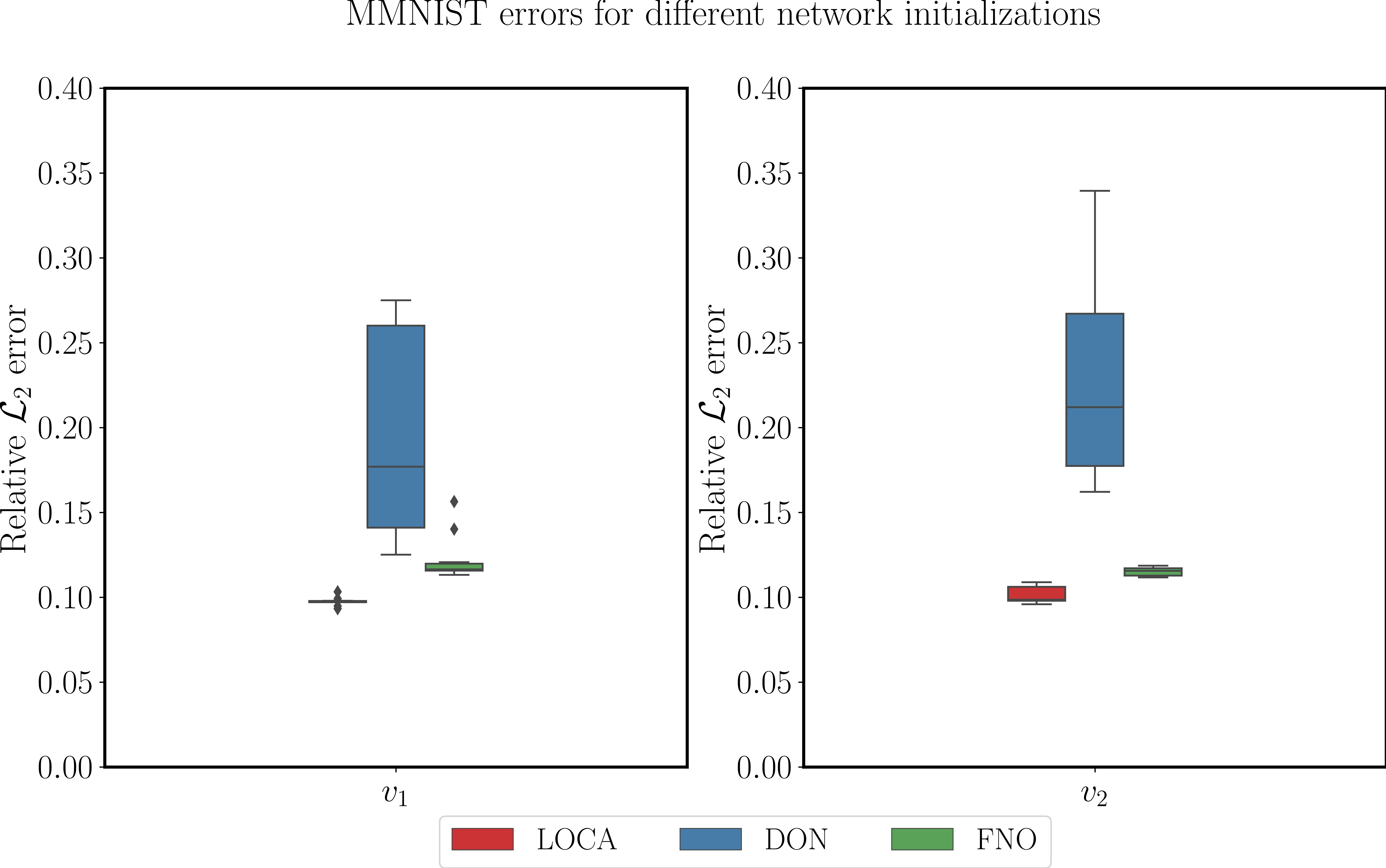}
\caption{(Robustness) Maximum relative $\mathcal{L}_2$ error boxplots for Mechanical MNIST with over random model initializations: The left and right subplots show the distribution of maximum errors over the testing data set for the horizontal and vertical displacements, respectively. We consider $7 \%$ of the available output function measurments for training and run the model for $10$ different random initializations. We observe that our method shows better performance than the other methods for both parameters $v_1$ and $v_2$.}
\label{fig:mmnsit_box_plot_errors_max}
\end{figure}

Next, we examine the variability of the models' performance with respect to the random initialization of the network parameters. We consider the Mechanical MNIST benchmark where the input data is clean but the output data contain noise. We train each model 10 times with different random seeds for initialization and record the maximum error in each case. In Figure \ref{fig:mmnsit_box_plot_errors_max} we present the distribution of maximum prediction errors under different random seeds for the displacement in horizontal and vertical directions, respectively. We observe that LOCA displays a smaller spread of error for the case of displacement in the horizontal direction, $v_1$, and similar performance to the FNO for the case of displacement in the vertical direction, $v_2$.

\begin{table}
\centering
\resizebox{1\textwidth}{!}{%
\begin{tabular}{|l|l|l|l|}
\hline
\backslashbox{Noise scenario and error }{Model}& FNO  & DON & LOCA \\ \hline
Clean to Clean (CC) & \bf{[0.004, 0.003]} & [0.028, 0.029] & [0.026, 0.026] \\ \hline
Clean to Noisy (CN) & [0.088, 0.087] & [0.061, 0.057] & \bf{[0.047, 0.049]} \\ \hline
Noisy to Noisy (NN) & \bf{[0.016, 0.016]} & [0.065, 0.064] & [0.032, 0.033] \\ \hline
Percentage error increase from CC to CN  & [1,930 $\%$, 2,238 $\%$] & [112$\%$, 96$\%$] & \bf{[80$\%$, 85$\%$]} \\ \hline
Percentage error increase from CC to NN  & [280 $\%$, 347$\%$]   & [128$\%$,120$\%$] & \bf{[26$\%$, 25$\%$]} \\ \hline
\end{tabular}
}
\caption{(Robustness)  Mechanical MNIST prediction error with noisy data: The first three rows present the mean relative $\mathcal{L}_2$ errors of the vertical and horizontal displacements $[ \text{Error}(v_1), \text{Error}(v_2)]$.  The last two rows show the percentage increase in mean error from the noiseless case to the scenarios where the testing input data is corrupted by noise and to the scenario that both the training and testing input data sets are corrupted by noise. For each case we consider $7 \%$ of the total data set as labeled data for training. We observe that our method shows the least percentage increase for each noise scenario.}

\label{tab:mmnist_percentage_increase}
\end{table}

\subsection{Generalization}
\label{sec:Generalization}

\begin{figure}
\centering
\includegraphics[width=.8\textwidth]{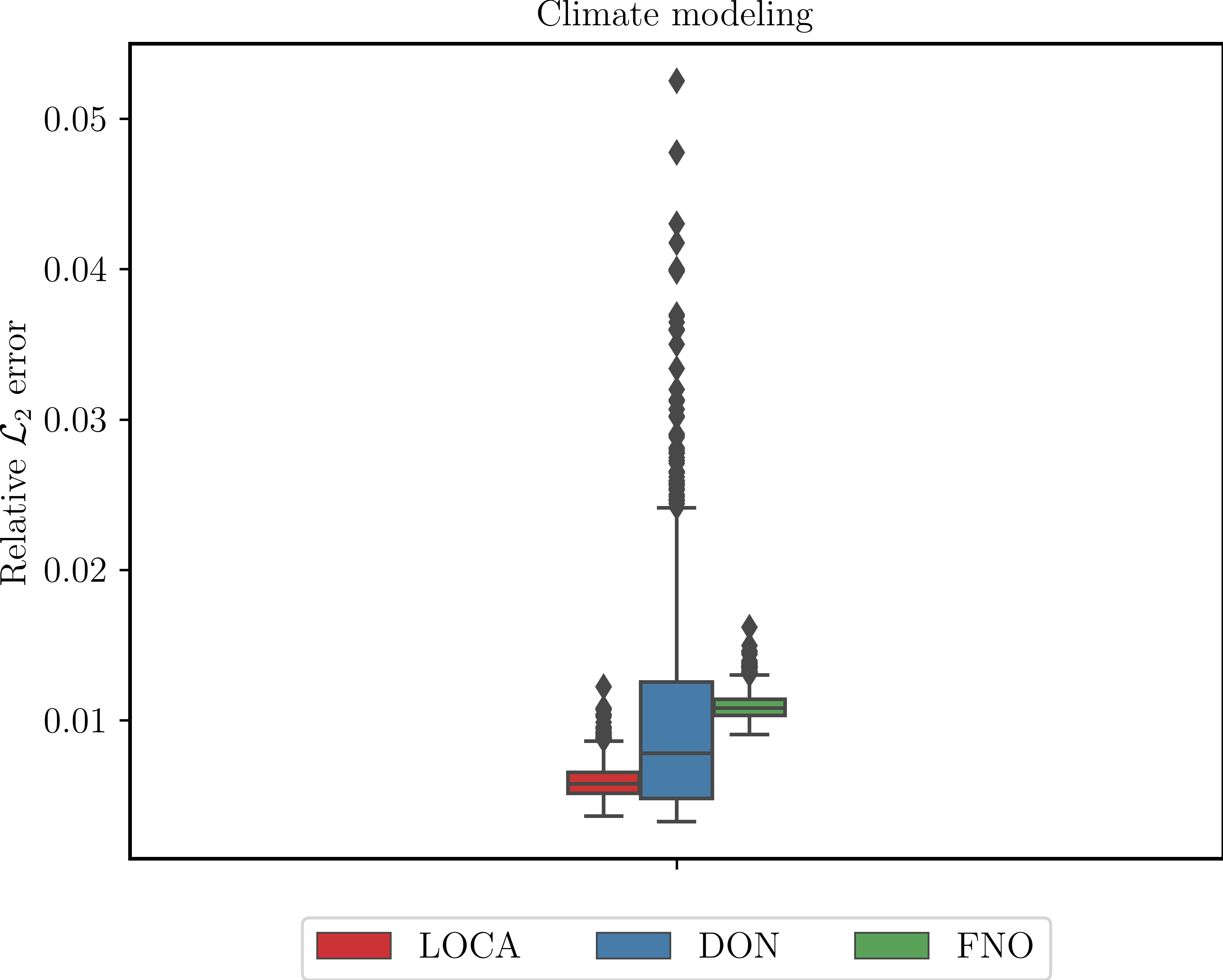}
\caption{(Generalization) Relative $\mathcal{L}_2$ error boxplots for the climate modeling experiment: We present the errors for the temperature prediction task on the testing data set. We consider $4 \%$ of the whole data set as labeled data used for training. We observe that our method performs better than the other methods both with respect to the median error and the error spread. }
\label{fig:weather_box_plot_errors}
\end{figure}

The ultimate goal of data-driven methods is to perform well outside of the data set they are trained on. This ability to generalize is essential for these models to be practically useful. In this section we investigate the ability of our model to generalize in three scenarios. We first consider an extrapolation problem where we predict the daily Earth surface air pressure from the daily surface air temperature. Our training data set consists of temperature and pressure measurements from 2000 to 2005 and our testing data set consists of measurements from 2005 to 2010. In Figure \ref{fig:weather_box_plot_errors}, we present the results for the extrapolation problem when considering $4\%$ of the available pressure measurements each day for training. We observe that our method achieves the lowest error rates while also maintaining a small spread of these errors across the testing data set. While the DON method achieves a competitive performance with respect to the median error, the error spread is larger than both LOCA and FNO with many outliers.

\begin{figure}
\centering
\includegraphics[width=.9\textwidth]{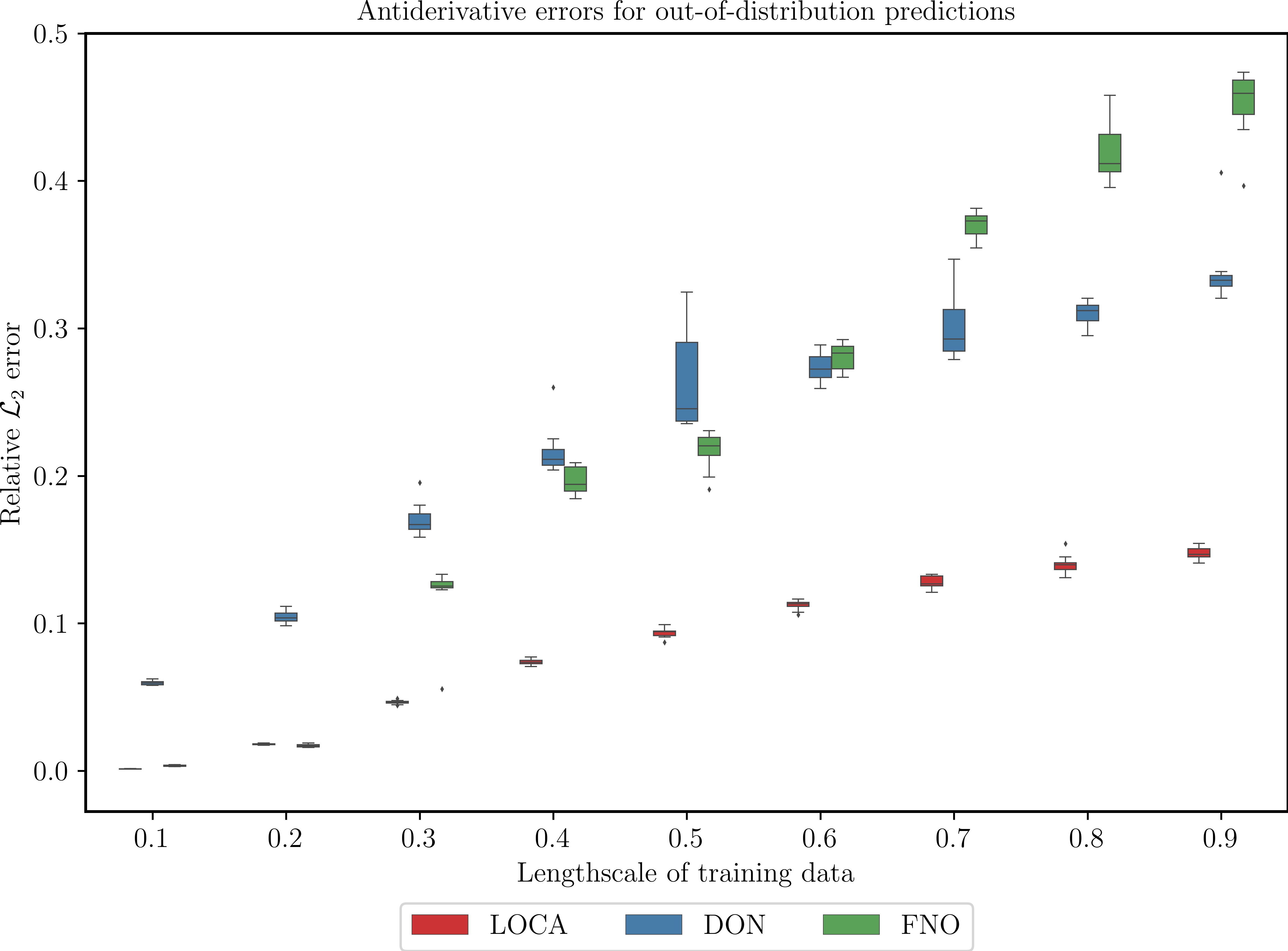}
\caption{(Generalization) Antiderivative relative $\mathcal{L}_2$ error boxplots for out-of distribution testing sets: We show the performance of all models when trained on increasingly out of distribution data sets from the testing set. We use all available output function measurements for training.}
\label{fig:antiderivative_box_plot_errors}
\end{figure}

Next, we examine the performance of our model under a distribution shift of the testing data. The goal of the experiment is to learn the antiderivatve operator where the training and testing data sets are sampled from a Gaussian process. We fix the length-scale of the testing distribution at $0.1$ and examine the effect of training over 9 different data sets with length-scales ranging from $0.1$ to $0.9$. In Figure \ref{fig:antiderivative_box_plot_errors}, we present the error on the testing data set after being trained on each different training data set. The error for each testing input is averaged over $10$ random network initializations. We observe that while the LOCA and FNO methods present a similar error for the first two cases, the FNO error is rapidly increasing. On the other hand, the DON method while presenting a larger error at first, eventually performs better than the FNO as the training length-scale increases. We find that LOCA outperforms its competitors for all cases.

\begin{figure}
\centering
\includegraphics[width=\textwidth]{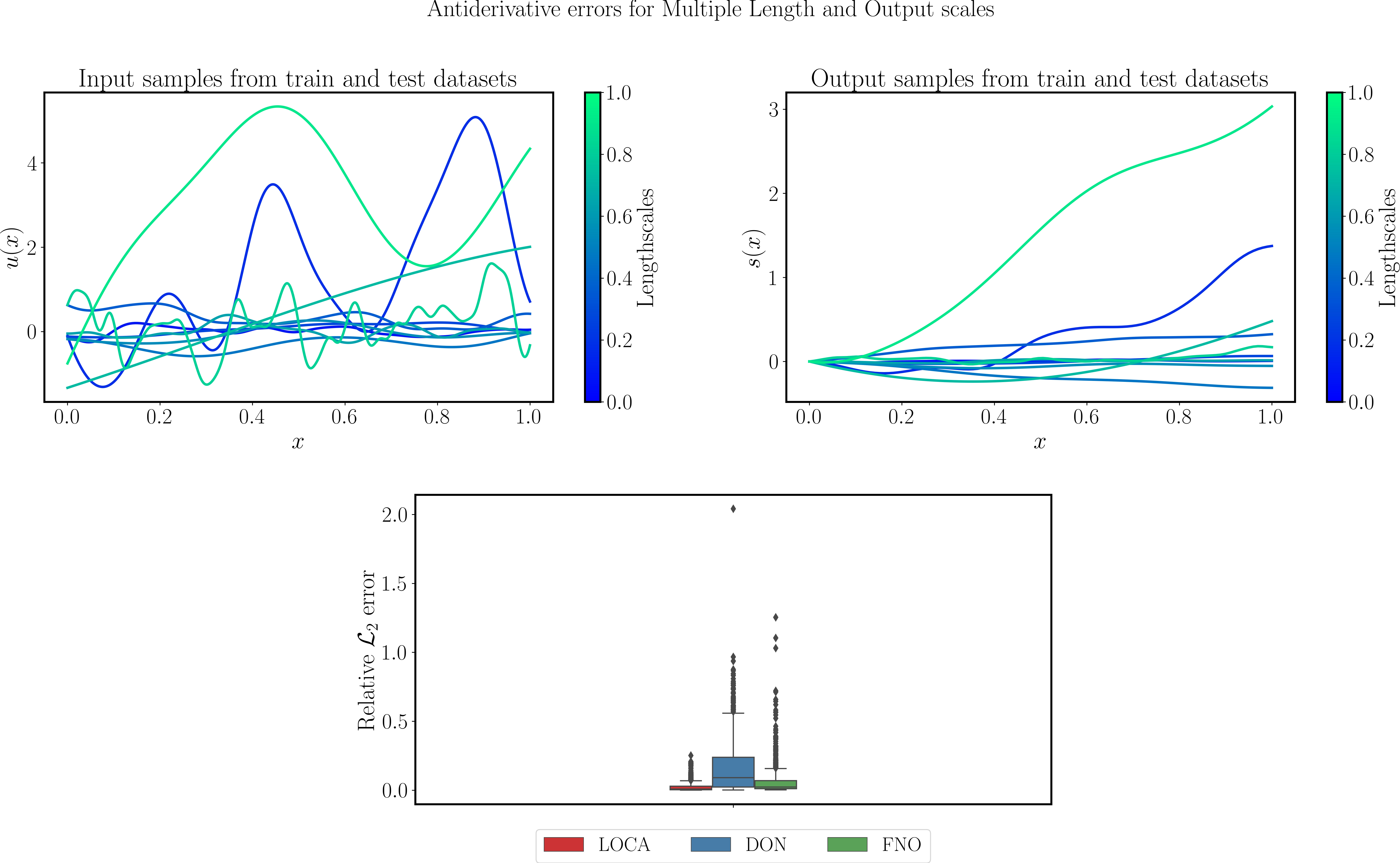}
\caption{(Generalization) Antiderivative relative $\mathcal{L}_2$ error boxplots given input functions with multiple lenghtscales and amplitudes: We present samples of the input and output functions from the testing data set in the top left and right figures, respectively, as well as the test error boxplots for each method, bottom figure.}
\label{fig:antiderivative_multi_scales_boxplot}
\end{figure}

Lastly, we examine the performance of the three models when the training and testing data set both contain a wide range of scale and frequency behaviors. We consider this set-up as a toy model for a multi-task learning scenario and we want to explore the generalization capabilities of our model for this case. We construct a training and testing data set by sampling inputs from a Gaussian process where the length-scale and amplitude are chosen over ranges of $2$ and $4$ orders of magnitude, respectively. In Figure \ref{fig:antiderivative_multi_scales_boxplot}, we present samples from the input distribution, the corresponding output functions, and the distribution of errors on the testing data set. We observe that our method is more accurate and the error spread is smaller than DON and Fourier Neural Operators. While the FNO method shows a median that is close to the LOCA model, there exist many outliers that reach very high error values.

\section{Discussion}\label{sec:discussion}

This work proposes a novel operator learning framework with approximation theoretic guarantees. Drawing inspiration from the Bahdanau attention mechanism \cite{Bahdanau2015}, the model is constructed by averaging a feature embedding of an input function over probability distributions that depend on the corresponding output function's query locations. 

A key novelty of our approach is the coupling of these probability distributions through a variation of the classic attention mechanism called {\em Kernel-Coupled Attention} (KCA). Instead of normalizing a single proposal score function $g$ defined over the query domain $\curlyY$, the KCA mechanism couples the score function across point in $\curlyY$ by integrating against a similarity kernel. Thus, the KCA mechanism is able to model correlations between different query scores explicitly instead of relying on the score function $g$ to learn these relations alone. We hypothesize, and support with experiments, that this property allows the model to learn very efficiently using a small fraction of labeled data. In order to have a feature encoder that is robust to small deformations and noise in the input, we employ a multi-resolution feature extraction method based on the wavelet scattering transform of Bruna \textit{et. al.} \cite{bruna2013invariant}. We empirically show that this is indeed a property of our model. Our experiments additionally show that the model is able to generalize across varying distributions of functional inputs, and is able to extrapolate on a functional regression task with global climate data.

A potential drawback of the proposed method is the computational cost needed for numerically approximating the integrals in the KCA mechanism. When using Monte-Carlo with $P$ query points there is a complexity of $O(P^2)$. Instead if a quadrature approach is taken with $P$ queries and $Q$ nodes there is a complexity of $O(PQ + Q^2)$. The relative efficiency of these two approaches in general will depend on the number of quadrature points necessary for a good approximation, and thus on the dimension of the query domain. In general, integrals over high dimensional domains will become increasingly costly to compute. 

Therefore, an immediate future research direction is to use further approximations to allow the kernel integral computations to scale to larger numbers of points and dimensions. A first approach is to parallelize this integral computation by partitioning the domain into pieces and summing the integral contributions from each piece. To lower the the computational complexity of the kernel computations between the query and integration points we can also use approximations of the kernel matrix. For example, in the seminal paper of Rahimi and Recht \cite{rahimi2007random} the authors propose an approximation of the kernel using a random feature strategy. More recently, in the context of transformer architectures, a number of approximations have been proposed to reduce the complexity of such computations to be linear in the number of kernel points $O(N)$. A non-exhaustive list of references include Linformers \cite{wang2020linformer}, Performers \cite{choromanski2020rethinking},  Nystr\"{o}formers \cite{xiong2021nystr} and Fast Transformers \cite{katharopoulos2020transformers}. 

Another potential extension of our framework is to take the output of our model as the input function of another LOCA module and thus make a layered version of the architecture. While in our experiments we did not see that this modification significantly increased the performance of the model, it is possible that other variants of this modular architecture could give performance improvements. Lastly, recall that the output of our model corresponds to the context vector generated in the Bahdanau attention. In the align and translate model of Bahdanau \textit{et. al.} \cite{Bahdanau2015} this context vector is used to construct a distribution over possible values at the output location. By using the output of our model as a context vector in a similar architecture, we can create a probabilistic model for the potential values of the output function, therefore providing a way to quantify the uncertainty associated with the predictions of our model.

A main application of operator learning methods is for PDEs, where they are used as surrogates for traditional numerical solvers. Since the forward pass of an operator learning model is significantly faster than classical numerical methods, the solution of a PDE under many different initial conditions can be expediently obtained. This can be a key enabler in design and optimal control problems, where many inputs must be tested in pursuit of identifying an optimal system configuration. A key advantage of operator learning techniques in this context is that they also allow the quick evaluation of sensitivities with respect to inputs (via automatic differentiation), thus enabling the use of gradient-based optimization. Conventional methods for computing sensitivities typically rely on solving an associated adjoint system with a numerical solver. In contrast, a well-trained operator learning architecture can compute these sensitivities at a fraction of the time. Therefore, we expect that successful application of operator learning methods to predict the output of physical systems from control inputs can have a significant impact in the design of optimal inputs and controls. Some preliminary work in this direction has been explored in Wang \textit{et. al.} \cite{wang2021fast}.

\section{Acknowledgements}
G..K. and P.P. would like to acknowledge support from the US Department of Energy under under the Advanced Scientific Computing Research program (grant DE-SC0019116), the US Air Force (grant AFOSR FA9550-20-1-0060), and US Department of Energy/Advanced Research Projects Agency (grant DE-AR0001201).  J.S. and G.P. would like to acknowledge support from the AFOSR under grant FA9550-19-1-0265 (Assured Autonomy in Contested Environments) and the NSF Simmons Mathematical and Scientific Foundations of Deep Learning (grant 2031985).  We also thank the developers of the software that enabled our research, including JAX \cite{jax2018github}, Kymatio \cite{andreux2020kymatio}, Matplotlib \cite{hunter2007matplotlib}, Pytorch \cite{paszke2019pytorch} and NumPy \cite{harris2020array}. We would also like to thank Andreas Kalogeropoulos and Alp Aydinoglu for their useful feedback on the manuscript.

\bibliographystyle{unsrt}
\bibliography{references}

\appendix
\section{Nomenclature}
Table \ref{tab:TableOfNotationForMyResearch} summarizes the main symbols and notation used in this work.

\begin{table}[t!]
\begin{center}
\begin{tabular}{r c p{10cm} }
\toprule
$[n]$ & & The set $\{1, \ldots, n\} \subset \N$. \\
$u \odot v$ & & Hadamard (element-wise) product of vectors $u$ and $v$. \\
$C(A,B)$ &  & Space of continuous functions from a space $A$ to a space $B$. \\
$C^1$ & & Space of continuous functions with continuous derivative. \\
$L^2$ & & Hilbert space of square integrable functions. \\
$\mathcal{H}_k$ & & Reproducing Kernel Hilbert Space with kernel $k$. \\
$\Delta^n$ & & $n$-dimensional simplex. \\
$\curlyX$ &  & Domain for input functions. \\
$\curlyY$ & & Domain for output functions.\\
$x$& & Input function arguments. \\
$y$&  & Output function arguments (queries). \\
$u$ & & Input function in $C(\curlyX, \R^{d_u})$. \\
$s$ & & Output function in $C(\curlyY, \R^{d_s})$. \\
$\mathcal{F}$&  & Operator mapping input functions $u$ to output functions $s$.\\
$g(y)$&  & Proposal score function.\\
$\tilde g(y)$ & & Kernel-Coupled score function. \\
$\varphi(y)$ & & Attention weights at query $y$. \\
$v(u)$&  & Feature encoder. \\
$\kappa(y,y')$&  & Coupling kernel.\\
$k(y,y')$&  & Base similarity kernel.\\
\bottomrule
\end{tabular}
\end{center}
\caption{(Nomenclature) A summary of the main symbols and notation used in this work.}
\label{tab:TableOfNotationForMyResearch}
\end{table}

\section{Proof of Theorem \ref{thm: universality}} \label{sec: main universality proof}

\begin{proof}
The starting point of the proof is the following lemma, which gives justification for approximating operators on compact sets with finite dimensional subspaces as in \cite{chen1995universal, lanthaler2021error, kovachkineural}.  The lemma follows immediately from the fact that for any compact subset $\curlyU$ of a Banach space $\curlyE$ and any $\epsilon > 0$, there exists a finite dimensional subspace $\curlyE_n \subset \curlyE$ such that $d(\curlyU, \curlyE_n) < \epsilon$.

\begin{lemma} \label{lem: lem 6}
Let $\curlyU \subset \curlyE$ be a compact subset of a Banach space $\curlyE$. Then for any $\epsilon > 0$, there exists $n \in \N$, $\phi_1, \ldots, \phi_n \in \curlyE$, and functionals $c_1, \ldots, c_n$ with $c_i: \curlyE \to \R$ such that
\begin{align*}
\sup_{u \in \curlyU} \|u - \sum_{i=1}^n c_i(u) \phi_i\|_\curlyE < \epsilon.
\end{align*}
\end{lemma}

Returning to the problem of learning a continuous operator $\curlyG: \curlyU \to C(\curlyY, \R^{d_s})$, since $\curlyU$ is assumed to be compact and $\curlyG$ is continuous, the image $\curlyG(\curlyU)$ is compact in the co-domain. Thus, we may apply Lemma \ref{lem: lem 6} to the set $\curlyG(\curlyU)$. This shows that for any $\epsilon > 0$, there exists $c_1, \ldots, c_n$ with each $c_i: \curlyU \to \R$ linear and continuous and functions $\phi_1, \ldots, \phi_n$ with $\phi_i: \curlyY \to \R^{d_s}$ such that
\begin{align} \label{eq: lem 6 operator}
\sup_{u \in \curlyU} \sup_{y \in \curlyY} \left\| \curlyG(u)(y) - \sum_{i=1}^n c_i(u) \phi_i(y)\right\| < \epsilon.
\end{align}

Next, we show that the approximation of $\curlyG(u)(y)$ given in \eqref{eq: lem 6 operator} can be expressed equivalently as vector of averages of a modified collection of functionals $c_i$. These functionals $c_i$ will form the coordinates of our input feature vector $v(u)$. First, for each $\phi_i:\curlyY \to \R^{d_s}$ we may form the positive and negative parts, whose coordinates are defined by
\begin{align*}
(\phi_i^+)_q &= \max\{(\phi_i)_q, 0\} \\
(\phi_i^-)_q &= - \min\{(\phi_i)_q, 0\}
\end{align*}
Note that $\phi_i^+$ and $\phi_i^-$ are continuous, non-negative, and that $\phi_i = \phi_i^+ - \phi_i^-$. For $j = 1, \ldots, 2n$ define a new collection of functions $\varphi_j: \curlyY \to \R^{d_s}$ by
\begin{align*}
\varphi_j = \begin{cases}
\frac{1}{2n \|\phi_i^+\|_\infty}\phi_i ^+ & \text{if } j = 2i \cr
\frac{1}{2n \|\phi_i^-\|_\infty}\phi_i^- & \text{if } j=2i-1\end{cases}
\end{align*}
and define
$$\varphi_{2n+1} := \mathbf{1}_{d_s} - \sum_{j=1}^{2n} \varphi_j.$$
By construction, for all $y$ we have that $\varphi_j(y) \in [0,1]^{d_s}$, 
\begin{align*} 
\mathrm{span}\{\phi_i\}_{i=1}^{n} \subseteq \mathrm{span}\{\varphi_j\}_{j=1}^{2n+1},
\end{align*}
and
\begin{align*} 
\sum_{j=1}^{2n+1} \varphi_j(y) = \mathbf{1}_{d_s}.
\end{align*}
In order to allow each output dimension of each $\varphi_j$ to have its own coordinate function, (and thus have $v(u) \in \R^{n \times d_s}$), for each $\varphi_j$, we create $d_s$ new functions,
\begin{align*}
    \varphi_{j,k}(y) := e_k \odot \varphi_{j}(y),
\end{align*}
where $e_k \in \R^{d_s}$ is the $k$-th standard basis vector in $\mathbb{R}^{d_s}$. Thus, we have constructed a collection of vectors $\varphi_{j,k}$ such that $\langle \varphi_{j,k}, e_m \rangle = 0$ if and only if $k \neq m$,
\begin{align*}
    \sum_{j=1}^{2n+1} \sum_{k=1}^{d_s} \varphi_{j,k}(y) = \mathbf{1}_{d_s}, \quad \forall y \in \curlyY,
\end{align*}
and
\begin{align*}
\mathrm{span}\{\phi_i\}_{i=1}^{n} \subseteq \mathrm{span}\big\{\varphi_{j,k}\big\}_{\substack{j \in [2n+1] \\ k \in [d_s]}} .
\end{align*}
Since from Lemma \ref{lem: lem 6} we know
$$d(\mathrm{span}\{\phi_i\}_{i=1}^{n}, \curlyG(\curlyU)) < \epsilon,$$
we conclude that 
$$d(\mathrm{span}\big\{\varphi_{j,k}\big\}_{\substack{j \in [2n+1] \\ k \in [d_s]}}, \curlyG(\curlyU)) < \epsilon,$$
and can conclude the statement of the theorem.
\end{proof}

\section{Proof of Proposition \ref{prop: phi univ}}
\begin{proof}
Note that $\mathrm{im}(T_k^{1/2}) = \curlyH_\kappa$, \cite{paulsen2016introduction}. Since $\kappa$ is universal, $\mathrm{im}(T_\kappa^{1/2}) = \curlyH_\kappa \subset C(\curlyY, \R^n)$ is dense. Thus, it suffices to show that $T_\kappa(\curlyA)$ is dense in $\mathrm{im}(T_\kappa^{1/2})$.
We will make use of the following fact, which we state as a lemma. 
\begin{lemma} \label{lem: dense lemma}
If $f: X \to Y$ is a continuous map and $A \subset X$ is dense, then $f(A)$ is dense in $\mathrm{im}(f)$.
\end{lemma}
By the above lemma, we have that $T_\kappa(\curlyA)$ is dense in $\mathrm{im}(T_\kappa)$. Now we must show that $\mathrm{im}(T_\kappa) \subset \mathrm{im}(T_\kappa^{1/2})$ is dense as well. This again follows from the above lemma by noting that $\mathrm{im}(T_k) = T_k^{1/2}(\mathrm{im}(T_k^{1/2}))$, and $\mathrm{im}(T_k^{1/2})$ is dense in the domain of $T_k^{1/2}$.
\end{proof}

\section{Proof of Proposition \ref{prop: PD symm}}\label{sec: symmetry PD proof}

In this section, we show that the coupling kernel is symmetric and positive semi-definite. These two conditions are necessary to obtain theoretical guarantees of universality. The symmetry of the kernel $\kappa$ follows immediately from the symmetry of the base kernel $k$ in \eqref{eq: RBF} and the form of $\kappa$ in \eqref{eq: kernel def}. To prove $\kappa$ is positive semi-definite we must show for any $v_1, \ldots, v_n \in \R$ and $y_1, \ldots, y_n \in \curlyY$,
\begin{align*}
    \sum_{i,j=1}^n v_i v_j \kappa(y_i, y_j) \geq 0.
\end{align*}
For ease of notation define
\begin{align*}
    Z_i := \left(\int_{\curlyY} k(q(y_i),q(z))\;dz\right)^{1/2}.
\end{align*}
Using the definition of $\kappa$ from \eqref{eq: kernel def},
\begin{align*}
    \sum_{i,j=1}^n v_i v_j \kappa(y_i, y_j) &= \sum_{i,j=1}^n v_i v_j \frac{k(q(y_i), q(y_j))}{Z_i Z_j} \\
    &= \sum_{i,j=1}^n \frac{v_i}{Z_i} \frac{v_j}{Z_j} k(q(y_i), q(y_j)) \\
    &\geq 0,
\end{align*}
where in the last line we have used the positive semi-definiteness of $k$.

Finally, the injectivity of the map $g$ would imply that the overall feature map of $\kappa$ is injective, which gives that the kernel is universal \cite{christmann2010universal}.

\section{Proof of Proposition \ref{prop: v univ}} \label{sec: v unif proof}
\begin{proof}
Since $\curlyU$ is compact, $h$ is uniformly continuous. Hence, there exists $\delta > 0$ such that for any $\|u - v\| < \delta$, $\|h(u) - h(v)\| < \epsilon/2$. Define $u_d := \sum_{i=1}^d \langle u, e_i \rangle e_i$. By the uniform convergence of $u_d \to u$ over $u \in \curlyU$, there exists $d$ such that for all $u \in \curlyU$, $\|u - u_d\| < \delta$. Thus, for all $u \in \curlyU$
\begin{align*}
    \|h(u) - h(u_d)\| < \frac{\epsilon}{2}.
\end{align*}
If we define $r: \R^d \to C(\curlyX, \R^{d_u})$ as
$$r(\alpha) := \sum_{i=1}^d \alpha_i e_i,$$
we may write $h(u_d) = (h\circ r )(\curlyD_d(u))$. Now, note that $h \circ r \in C(\R^d, \R^n)$, and recall that, by assumption, the function class $\curlyA_d$ is dense in $C(\R^d, \R^n)$. This means there exists $f \in \curlyA_d$ such that $\|f - h \circ r\| < \epsilon/2$. Putting everything together, we see that
\begin{align*}
    \|h(u) - f \circ \curlyD_d(u)\| \leq \|h(u) - h(u_d)\| + \|(h \circ r)(\curlyD_d(u)) - f \circ \curlyD_d(u)\| < \epsilon.
\end{align*}
\end{proof}

\section{Supplementary Information for Experiments}
\label{sec:implementation_details}

In this section, we present supplementary information on the experiments presented in Section \ref{sec:Results}. 

\subsection{Computational Complexity}
\label{sec:Computational Complexity}
In LOCA the most expensive operations are the integral computations in the KCA mechanism. Let $z_1, \ldots, z_Q$ be the integration nodes with $z = [z_1, \ldots, z_Q]^\top$, and let the associated weights be $w_1, \ldots, w_Q$, with $w = [w_1, \ldots, w_Q]^\top$. For evaluating the KCA mechanism at a single query location $y_0$ with $Q$ integration nodes we are required to compute the matrices $k(y_0,z) \in \R^{1 \times Q}$, with $[k(y_0,z)]_{j} = k(y_0,z_j)$ and $k(z,z) \in \R^{Q \times Q}$, with $[k(z,z)]_{ij} = k(z_i, z_j)$. These are combined to compute the kernel $\kappa$ as
\begin{align*}
    \kappa(y_0, z) \approx \frac{1}{\left(k(y_0,z)w\right)^{1/2}}k(y_0,z) \odot \left(k(z,z) w \right)^{-1/2},
\end{align*}
where the exponent of $-1/2$ in the last factor is applied coordinate-wise. Carrying out this computation requires $Q$ steps to compute $k(y_0,z)$ and $Q^2$ steps to compute $k(z,z)$, giving an overall complexity of $Q + Q^2$. When considering $P>1$ the complexity for computing $k(y_0,z)$ becomes $PQ$ and the overall complexity becomes $PQ + Q^2$ because we need to compute $k(z,z)$ only once. For the Monte Carlo case, $Q=P$ and $y = z$, so we only need to make one computation of $k(y,y)$.  Therefore in this case, we have a complexity of $Q^2$.

Both methods have their benefits and disadvantages: in the Monte Carlo case, we need to perform $P^2$ computations once, but the cost scales exponentially with $P$. On the other hand, Gauss-Legendre quadrature requires $PQ + Q^2$ evaluations, but if $Q$ is small the overall computational cost is less than Monte-Carlo integration.

\subsection{Architecture Choices and Hyper-parameter Settings}
\label{sec:training_details}

In this section we present the neural network architecture choices, the training details, the training wall-clock time, as well as the number of training parameters for each model compared in the experiments. Specifically, for the DON and FNO models, we have performed an extensive number of simulations to identify settings for which these competing methods achieve their best performance.

For LOCA and DON we set the batch size to be $100$ and use exponential learning rate decay with a decay-rate of 0.99 every $100$ training iterations. For the FNO training, we set the batch size to be $100$ and consider a learning rate $l_r = 0.001$, which we then reduce by $0.5$ every $100$ epochs and a weight decay of $0.0001$. Moreover, for the FNO method we use the ReLU activation function.

\subsubsection{LOCA} For the LOCA model, we present the structure of the functions $g$, $f$, and $q$ in Table \ref{tab:architecture_choices_LOCA}. In Table \ref{tab:example_choices_LOCA} we present the number of samples considered for the train and test data sets, the number of points where the input and the output functions are evaluated, the dimensionality of positional encoding, the dimensionality of the latent space where we evaluate the expectation $\mathbb{E}(u)(y)$, the batch size used for training, and the number of training iterations. We present the parameters of the wavelet scattering network in Table \ref{tab:scatteringChoice}. The method used for computing the kernel integral for each example is presented in Table \ref{tab:integral_computation}.

\subsubsection{DON} For the DON model, we present the structure of $b$ and $t$, the branch and the trunk functions, in Table \ref{tab:architecture_choices_DON}. In Table \ref{tab:example_choices_DON} we present the number of samples considered for the train and test data sets, the number of points where the input and the output functions are evaluated, the dimensionality of the positional encoding, the dimensionality of the latent space, the batch size used for training, and the number of training iterations. In order to achieve competitive performance, we also adopted some of the improvements proposed in Lu \textit{et. al.} \cite{lu2021comprehensive}, including the application of harmonic feature expansions to both input and outputs, as well as normalization of the output functions. 

\subsubsection{FNO} For the FNO model, we present the architecture choice in Table \ref{tab:architecture_choices_FNO}. In Table \ref{tab:example_choices_FNO} we present the number of samples considered for the train and test data sets, the number of points where the input and the output functions are evaluated, the batch size used for training and the number of training epochs. 

\begin{table}
\centering
\resizebox{0.75\textwidth}{!}{%
\begin{tabular}{|c|c|c|c|c|c|c|}
\hline
Example & $g$ depth & $g$ width  & $f$ depth & $f$ depth & $q$ depth & $q$ width\\ 
\hline
 Antiderivative & 2  & 100 & 1 & 500 & 2 & 100\\
\hline
 Darcy Flow & 2  & 100 &2 & 100 & 2 & 100\\
 \hline
 Mechanical MNIST & 2 & 256 & 2 & 256& 2 & 256\\
\hline
 Shallow Water Eq. &  1 & 1024 & 1 & 1024 & 1 & 1024\\
\hline
 Climate Modeling & 2 & 100 & 2 & 100&  2 & 100\\
\hline
\end{tabular}
}
\caption{LOCA Architectural choices for each benchmark considered in this work: We present the chosen architecture for $g$ and $q$, the functions that constructs $\phi(y)$, and the function $v$ which together build up the architecture of the LOCA model. }
\label{tab:architecture_choices_LOCA}
\end{table}
 
\begin{table}
\centering
\resizebox{0.97\textwidth}{!}{%
\begin{tabular}{|c|c|c|c|c|c|c|c|c|c|}
\hline
Example & $N_{train}$ & $N_{test}$ & m & P & $n$ & $H$ & $l$ & Batch $\#$& $\#$ of train iterations \\
\hline
 Antiderivative & 1000  & 1000 & 100 & 100 & 100 & 10 & 100 & 100 & 50000 \\
\hline
 Darcy Flow & 1000  & 1000 & 1024 & - & 100 & 6 & 100 & 100 & 20000 \\
 \hline
 Mechanical MNIST &  60000 & 10000 & 784 & 56 & 500 & 10 & 100 & 100 &100000\\
\hline
 Shallow Water Eq. &  1000 & 1000 & 1024 & 128 & 480 & 2 & 100& 100& 80000\\
\hline
 Climate Modeling &  1825 & 1825 & 5184 & 144 & 100 & 10 & 100& 100 & 100000\\
\hline
\end{tabular}
}
\caption{LOCA model parameters for each benchmark considered in this work: We present the numbers of training and testing data $N_{train}$ and $N_{test}$, respectively, the number of input coordinate points $m$ where the input function is evaluated, the number of coordinates $P$ where the output function is evaluated, the dimension of the latent space $n$ over which we evaluate the expectation, the number of positional encoding features $H$ for the positional encoding, the dimensionality of the encoder $l$, the size of the batch $B$ and the iterations for which we train the model.}
\label{tab:example_choices_LOCA}
\end{table}

\begin{table}
\centering
\resizebox{0.33\textwidth}{!}{%
\begin{tabular}{|c|c|c|c|}
\hline
Example & $J$& $L$ & $m_o$ \\ 
\hline
 Antiderivative & 4  &  8 & 2\\
\hline
 Darcy Flow &1  &  2 & 2\\
 \hline
 Mechanical MNIST & 1 & 16 & 2\\
\hline
 Shallow Water Eq. &  1 & 8 & 2\\
\hline
 Climate Modeling & 1 & 8 & 2\\
\hline
\end{tabular}
}
\caption{Chosen parameters for the wavelet scattering network: $J$ represents the log-2 scatteting scales, $L$ the angles used for the wavelet transform and $m_o$ the maximum order of scattering coefficients to compute. The wavelet scattering network is implemented using the Kymatio library \cite{andreux2020kymatio}.}
\label{tab:scatteringChoice}
\end{table}

\begin{table}
\centering
\resizebox{0.39\textwidth}{!}{%
\begin{tabular}{|c|c|}
\hline
Example &  Integration method \\ 
\hline
 Antiderivative & Quadrature \\
\hline
 Darcy Flow &  Quadrature \\
 \hline
 Mechanical MNIST  & Quadrature \\
\hline
 Shallow Water Eq. & Monte Carlo \\
\hline
 Climate Modeling &  Monte Carlo \\
\hline
\end{tabular}
}
\caption{Integral computation method for each benchmark considered in this work: We present the method that is used to compute the required kernel integrals for the LOCA method.}
\label{tab:integral_computation}
\end{table}


\begin{table}
\centering
\resizebox{0.55\textwidth}{!}{%
\begin{tabular}{|c|c|c|c|c|}
\hline
Example & $b$ depth & $b$ width  & $t$ depth & $t$ depth \\ 
\hline
 Antiderivative & 2  & 512 & 2 & 512 \\
\hline
 Darcy Flow & 6  & 100 & 6 & 100 \\
 \hline
 Mechanical MNIST & 4 & 100 & 4 & 100 \\
\hline
 Shallow Water Eq. &  11 & 100 & 11 & 100 \\
\hline
 Climate Modeling & 4 & 100 & 4 & 100 \\
\hline
\end{tabular}
}
\caption{DON architecture choices for each benchmark considered in this work.}
\label{tab:architecture_choices_DON}
\end{table}
 
\begin{table}
\centering
\resizebox{0.97\textwidth}{!}{%
\begin{tabular}{|c|c|c|c|c|c|c|c|c|c|}
\hline
Example & $N_{train}$ & $N_{test}$ & m & P & $n$ & $H$ & $l$ & Batch $\#$& Train iterations \\
\hline
 Antiderivative & 1000  & 1000 & 1000 & 100 & 100 & 2 & 100 & 100 & 50000 \\
\hline
 Darcy Flow & 1000  & 1000 & 1024 & - & 100 & 6 & 100 & 100 & 20000 \\
 \hline
 Mechanical MNIST &  60000 & 10000 & 784 & 56 & 500 & 10 & 100 & 100 &100000\\
\hline
 Shallow Water Eq. &  1000 & 1000 & 1024 & 128 & 480 & 2 & 100& 100& 80000\\
\hline
 Climate Modeling &  1825 & 1825 & 5184 & 144 & 100 & 10 & 100& 100 & 100000\\
\hline
\end{tabular}
}
\caption{DON model parameter for each benchmark considered in this work: We present the numbers of training and testing data $N_{train}$ and $N_{test}$, respectively, the number of input coordinate points $m$ where the input function is evaluated, the number of coordinates $P$ where the output function is evaluated, the dimension of the latent space $n$ over which we evaluate the inner product of the branch and the trunk networks, the number of positional encoding features $H$, the size of the batch $B$ and the iterations for which we train the model.}
\label{tab:example_choices_DON}
\end{table}


\begin{table}
\centering
\resizebox{0.6\textwidth}{!}{%
\begin{tabular}{|c|c|c|c|}
\hline
Example & $\#$ of modes & width  & $\#$ of FNO layers \\
\hline
 Antiderivative & 32  & 100 & 4 \\
\hline
 Darcy Flow & 8  & 32 & 4 \\
 \hline
 Mechanical MNIST & 12 & 32 & 4 \\
\hline
 Shallow Water Eq. &  8 & 25 & 4 \\
\hline
 Climate Modeling & 12 & 32 & 4 \\
\hline
\end{tabular}
}
\caption{FNO architecture choices for each benchmark considered in this work.}
\label{tab:architecture_choices_FNO}
\end{table}
 
\begin{table}
\centering
\resizebox{0.77\textwidth}{!}{%
\begin{tabular}{|c|c|c|c|c|c|c|c|c|}
\hline
Example & $N_{train}$ & $N_{test}$ & m & P & Batch $\#$& Train Epochs \\
\hline
 Antiderivative & 1000  & 1000 & 1000 & 100 &  100 & 500 \\
\hline
 Darcy Flow & 1000  & 1000 & 1024 & - &  100 & 500 \\
 \hline
 Mechanical MNIST &  60000 & 10000 & 784 & 56 &  100 & 200\\
\hline
 Shallow Water Eq. &  1000 & 1000 & 1024 & 128 & 100 & 400\\
\hline
 Climate Modeling &  1825 & 1825 & 5184 & 144 & 73 & 250\\
\hline
\end{tabular}
}
\caption{FNO model parameter for each benchmark considered in this work: We present the numbers of training and testing data $N_{train}$ and $N_{test}$, respectively, the number of input coordinate points $m$ where the input function is evaluated, the number of coordinates $P$ where the output function is evaluated, the size of the batch $B$ and the epochs for which we train the model.}
\label{tab:example_choices_FNO}
\end{table}

\begin{table}
\centering
\resizebox{0.55\textwidth}{!}{%
\begin{tabular}{|c|c|c|c|}
\hline
Example & LOCA & DON & FNO \\
\hline
 Antiderivative & 1,677,300 & 2,186,672 &  1,333,757 \\
\hline
 Darcy Flow & 381,000  & 449,400  & 532,993  \\
 \hline
 Mechanical MNIST & 2,475,060  & 3,050,300 & 1,188,514  \\
\hline
 Shallow Water Eq. & 5,528,484  & 5,565,660 & 5,126,690 \\
\hline
 Climate Modeling & 1,239,500 & 5,805,800 & 1,188,353 \\
\hline
\end{tabular}
}
\caption{Total number of trainable parameters for each model, and for each benchmark considered in this work.}
\label{tab:number_of_parameters}
\end{table}

\subsection{Computational Cost}
\label{sec:Computational Cost}

We present the wall clock time, in minutes, needed for training each model for each different example presented in the manuscript in Table \ref{tab:computational_cost}. For the case of the Darcy flow, the computational time is calculated for the case of $P=1024$, meaning we use all available labeled output function measurements per training example. We choose this number of query points to show that even when the number of labeled data is large, the computational cost is still reasonable, despite the KCA computation bottleneck. We observe that the wall clock time for all methods lie in the same order of magnitude. All the models are trained on a single NVIDIA RTX A6000 GPU. 

\begin{table}
\centering
\resizebox{0.50\textwidth}{!}{%
\begin{tabular}{|c|c|c|c|}
\hline
Example & LOCA & DON & FNO \\
\hline
 Antiderivative & 2.23 & 2.08 &  2.06 \\
 \hline
 Darcy Flow ($P=1024$) & 5.51  & 3.5  & 1.50  \\
 \hline
 Mechanical MNIST & 21.70  & 16.61 & 22.87  \\
\hline
 Shallow Water Eq. & 12.10  & 15.39 & 13.95 \\
\hline
 Climate Modeling & 4.52 & 7.51 & 10.49 \\
\hline
\end{tabular}
}
\caption{Computational cost for training each model across all benchmarks considered in this work: We present the wall clock time \emph{in minutes} that is needed to train each model on a single NVIDIA RTX A6000 GPU.}
\label{tab:computational_cost}
\end{table}

\subsection{Comparison Metrics}
\label{sec:comparison_metrics}

Throughout this work, we employ the relative $\curlyL_2$ error as a metric to assess the test accuracy of each model, namely:
\begin{equation*}
    \text{Test error metric} =  \frac{|| s^i(y) - \hat{s}^i(y) ||^2_2}{|| s^i(y)||^2_2},
\end{equation*}
where $\hat{s}(y)$ the model predicted solution, $s(y)$ the ground truth solution and $i$ the realization index. The relative $\curlyL_2$ error is computed across all examples in the testing data set, and different statistics of the error vector are computed: the median, quantiles, and outliers. For all examples the errors are computed between the full resolution reconstruction and the full resolution ground truth solution.

\subsection{Experiments} \label{sec:APexperiments}

In this section, we present additional details about the experimental scenarios discussed in Section \ref{sec:Results}.

\subsubsection{Antiderivative}

We approximate the antiderivative operator for demonstrating the generalization capabilities of the LOCA in two inference scenarios. The antiderivative operator is defined as
\begin{equation*}
    \frac{ds(x)}{dx} = u(x), \quad s(x) = s_0 + \int_{0}^x u(\tau) d\tau
    \label{equ:antiderivative},
\end{equation*}
where we consider $x \in \curlyX = [0,1]$ and the initial condition $s(0)=0$. For a given forcing term $u$ the solution operator $\curlyG$ of system \eqref{equ:antiderivative} returns the antiderivative $s(x)$. In the notation of our model, the input and output function domains coincide, $\curlyX = \curlyY$ with $d_x = d_y = 1$. Since  the solution operator is a map between scalar functions, we also have $d_u = d_s = 1$. Under this setup, our goal is to learn the solution operator $\curlyG: C(\curlyX, \R) \to C(\curlyX, \R)$. 

To construct the data sets we sample the forcing function $u(x)$ from a Gaussian process prior and measure these functions at $500$ points. We numerically integrate them to obtain $100$ measurements of each output function to use for training different operator learning models. 

For investigating the performance of LOCA on out-of-distribution prediction tasks, we create  training data sets by choosing $l_{train} \in [0.1, 0.9]$, and consider $9$ cases of increasing $l_{train}$ spaced by $0.1$ each. The training and testing data sets each have $N=1,000$ solutions of equation \eqref{equ:antiderivative}, and we use  $100 \%$ of all available output evaluation function points, both for training and testing.

For the case where we train and test on multiple length and output scales, we construct each example in the data set as follows. To construct each input sample, we first we sample a uniform random variable $\delta \sim \mathcal{U}(-2,1)$, and set the corresponding input sample length-scale to $l = 10^{\delta}$. Similarly, we construct a random amplitude scale by sampling $\zeta \sim \mathcal{U}(-2,2)$, and setting $o = 10^{\zeta}$. Then we sample $u(x)$ from a Gaussian Process prior $u(x) \sim GP(0, \text{Cov}(x,x'))$, where $\text{Cov}(x, x') = o \exp \Big ( {\frac{\| x- x' \|}{ l }} \Big )$. The length and the outputs scales are different for each realization, therefore we have $1,000$ different length and outputs scales in the problem.

\subsubsection{Darcy Flow}

Fluid flow through porous media is governed by Darcy's Law \cite{bear2013dynamics}, which can be mathematically expressed by the following partial differential equation system,
\begin{equation} \label{eq: darcy}
\begin{split}
        & \nabla \cdot ( u(x) \nabla s(x)) = f(x), \quad x \in \curlyX,
\end{split}
\end{equation}
subject to appropriate boundary conditions
\begin{alignat*}{2}
        s &= 0  && \quad \text{ on } \Gamma_\curlyX, \\
        (u(x)  \nabla  s(x)) \cdot n &= g && \quad \text{ on } \Gamma_N, 
\end{alignat*}{2}
where $u$ is permeability of the porous medium, and $s$ is the corresponding fluid pressure. Here we consider a domain $\curlyX = [0,1] \times [0,1]$ with a Dirichlet boundary  $\Gamma_D = \{ (0,x) \cup (1,x)\;|\;x_2 \in [0,1] \subset \partial \curlyX \}$, and a Neumann boundary $\Gamma_N = \{ (x,0) \cup (x,1) \;|\;x \in [0,1]\subset \partial \curlyX \}$. 

For a given forcing term $f$ and set of boundary conditions, the solution operator $\curlyG$ of system \eqref{eq: darcy} maps the permeability function $u(x)$ to the fluid pressure function $s(x)$. In the notation of our model, the input and output function domains coincide, $\curlyX = \curlyY$ with $d_x = d_y = 2$. Since in this case the solution operator is a map between scalar functions, we also have $d_u = d_s = 1$. Under this setup, our goal is to learn the solution operator $\curlyG: C(\curlyX, \R) \to C(\curlyX, \R)$. 

We set the Neumann boundary condition to be $g(x) = \sin(5x)$, the forcing term $f(x) = 5 \exp( - ((x_1-0.5)^2 + (x_2-0.5)^2))$, and sample the permeability function $u(x)$ from a Gaussian measure, as $u(x) = \exp(u_0 \cos(x))$ with $u_0 \sim \curlyN(0, 7^{3/2}(- \Delta + 49 I)^{-1.5}$. The training and testing data sets are constructed by sampling the initial condition along a $32 \times 32$ grid and solving the forward problem with the Finite Element library, Fenics \cite{alnaes2015fenics}. This gives us access to $32 \times 32$ solution values to use for training different operator learning models. Sub-sampling these solution values in the manner described in Section \ref{sec:Results} allows us to create training data sets to examine the effect of using only a certain percentage of the available data.

\begin{figure}
\centering
\includegraphics[width=\textwidth]{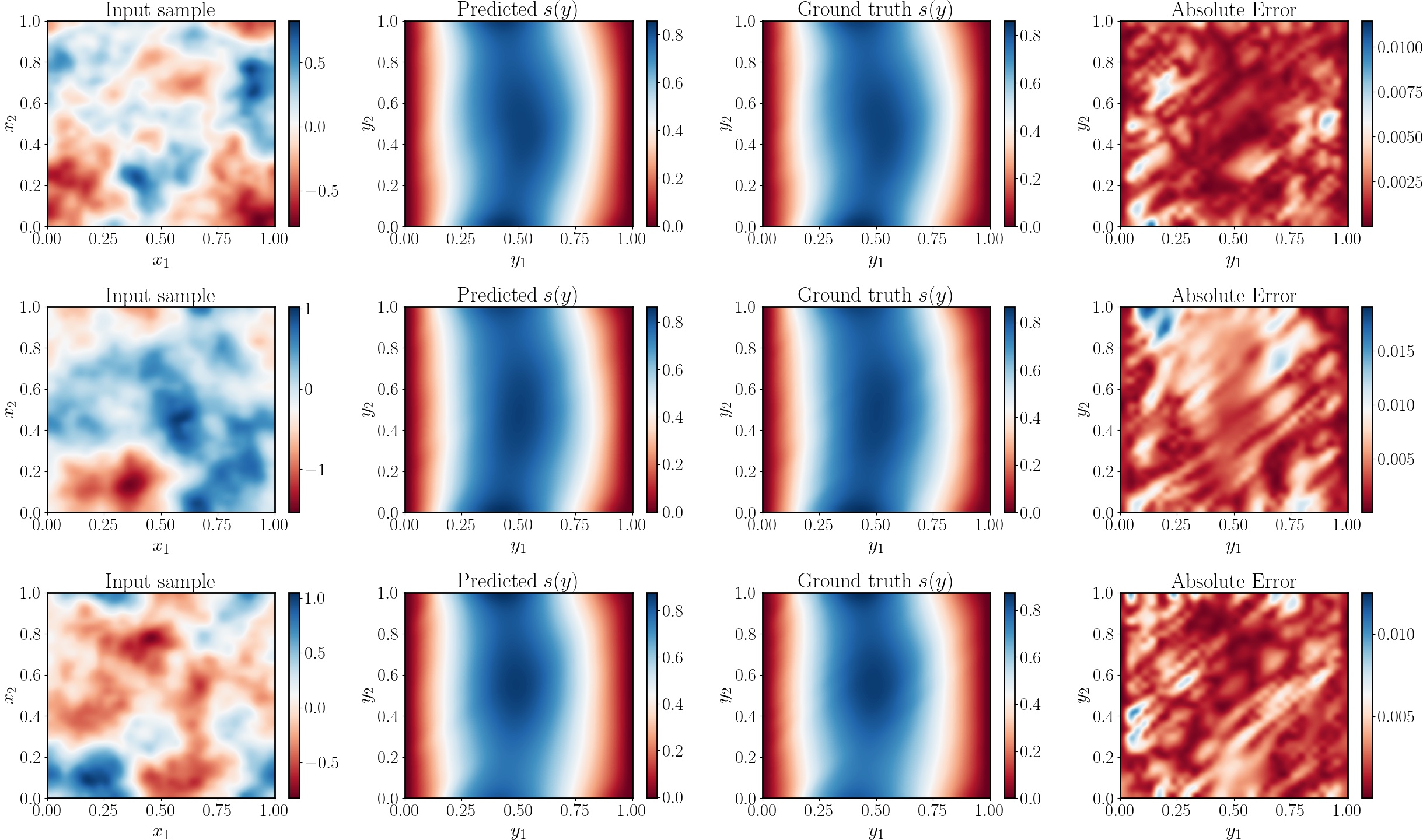}
\caption{Comparison between the full resolution prediction and ground truth for the flow through porous medium data set: We present the input sample, the prediction, the ground truth  and the absolute error for three realizations of the Darcy flow system.}
\label{fig:Darcy_solution_both_testmin_all}
\end{figure}

Figure \ref{fig:Darcy_solution_both_testmin_all} gives a visual comparison of the outputs of our trained model against the ground truth for three randomly chosen initial conditions, along with a plot of the point-wise error. We see that our model performs well across random initial conditions that were not present in the training data set. 

\subsubsection{Mechanical MNIST}
\label{sec:APMMNIST}

For this example, our goal is to learn the operator that maps initial deformations to later-time deformations in the equi-biaxial extension benchmark from the Mechanical MNIST database \cite{lejeune2020mechanical}. The data set is constructed from the results of $70,000$ finite-element simulations of a heterogeneous material subject to large deformations. MNIST images are considered to define a heterogeneous block of material described by a compressible Neo-Hookean model \cite{lejeune2020mechanical}. 

In our case, we are interested in learning displacement fields at later times, given some initial displacement. The material constitutive law is described by Lejeune \textit{et. al.} \cite{lejeune2020mechanical}

\begin{equation}
\label{equ:block material model}
    \psi = \frac{1}{2} \mu \Big [ \mathbf{F} :  \mathbf{F} - 3 -2\text{ln}(\text{det} \mathbf{F}) \Big]  + \frac{1}{2} \lambda \Big [ ((\text{det} \mathbf{F})^2 -1 ) - \text{ln} (\text{det} \mathbf{F}) \Big], 
\end{equation}
where $\psi$ is the strain energy, $\mathbf{F}$ is the deformation energy, $\mu$ and $\lambda$ are Lam{\'e} constants that can be found from the Young's modulus and the Poisson ratio
\begin{equation*}
    E = \frac{\mu ( 3 \lambda + 2 \mu)}{\lambda + \mu}, \quad \quad \nu = \frac{\lambda}{2 ( \lambda + \mu)}.
\end{equation*}
The Young's modulus is chosen based on the bitmap values to convert the image to a material as
\begin{equation*}
    E = \frac{b}{255} (100 - 1) + 1,
\end{equation*}
where $b$ is the bitmap value. Here, the Poisson ratio is fixed to $\nu = 0.3$ for all block materials. This means that the pixels inside the digits are block materials that are much stiffer than the pixels that are outside of the digits. For the equi-biaxial extension experiments, Dirichlet boundary conditions are applied by considering different displacement values $\mathbf{d} = [ 0.0, 0.001, 0.01, 0.1, 0.5 , 1 , 2, 4, 6, 8, 10, 12, 14]$ for the right and top of the domain, and $-\mathbf{d}$ for the left and bottom of the domain. 

In this benchmark the input and the output function domains  coincide, $\curlyX = \curlyY$ with $d_x = d_y = 2$, while the solution operator, $\curlyG$, is a map between vector fields with $d_u = d_s = 2$. Consequently, our goal here is to learn the solution operator $\curlyG: \curlyC(\curlyX, \mathbb{R}^2) \mapsto \curlyC(\curlyX, \mathbb{R}^2)$. Even though we create a map between displacement vectors, we present the magnitude of the displacement
\begin{equation*}
    s = \sqrt{v_1^2 + v_2^2},
\end{equation*}
for visual clarity of our plots. 

The data set is constructed by sampling MNIST digits on a $28 \times 28$ grid and solving equation \ref{equ:block material model} using the Finite Element library, Fenics \cite{alnaes2015fenics}. Out of the $70,000$ realizations that the MNIST data set contains, $60,000$ are used for training and $10,000$ are used for testing, therefore $N_{train} = 60,000$ and $N_{test} = 10,000$. We randomly sub-sample the number of measurement points per output function, as explained in Section \ref{sec:Results}, to create a training data set to demonstrate that our model only needs a small amount of labeled data to provide accurate predictions.

\begin{figure}
\centering
\includegraphics[width=1\textwidth]{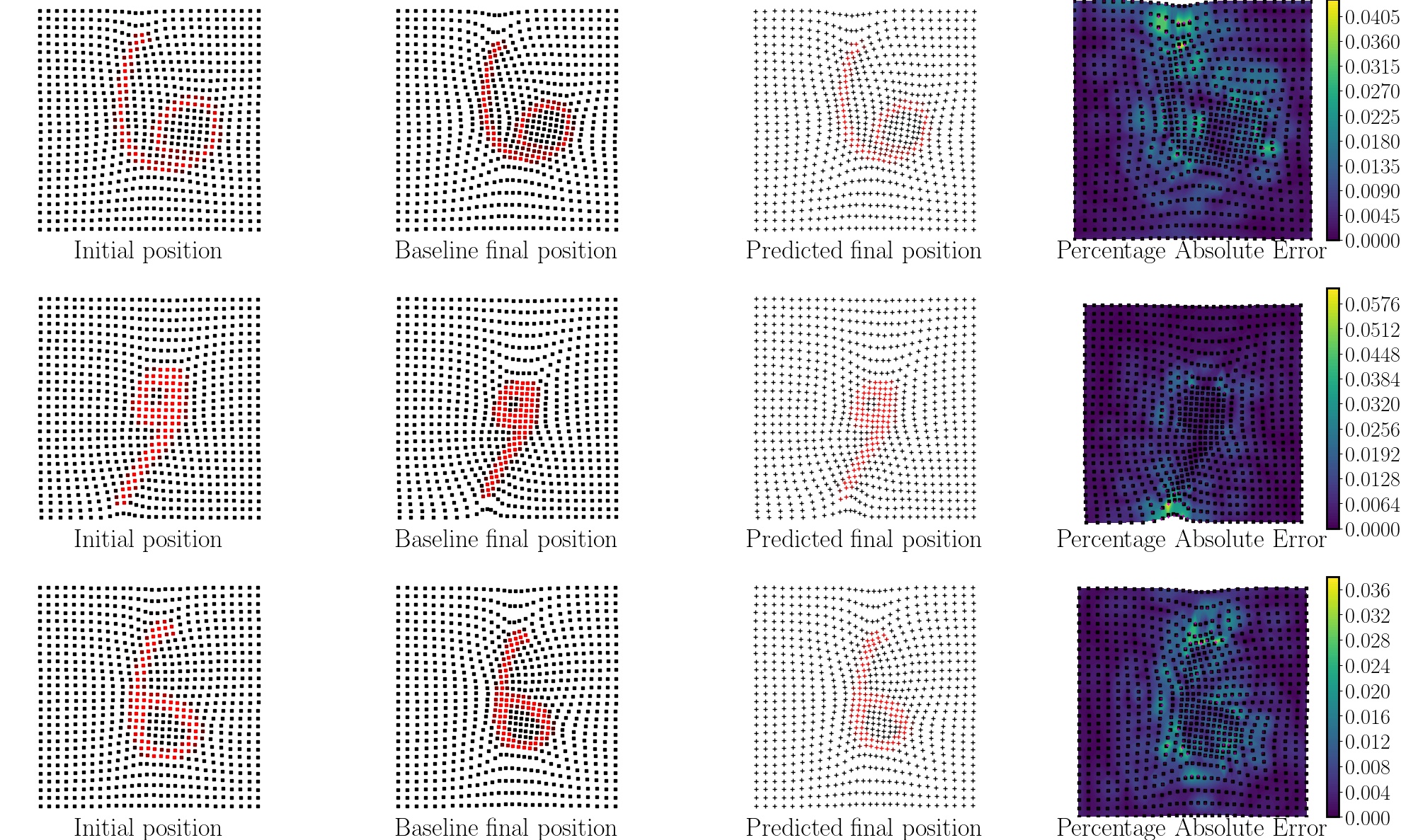}
\caption{Comparison between the predicted and the ground truth displacement magnitudes of the Mechanical MNIST case: We present the results of our model for 3 different MNIST digits under final displacement $d=14$. Despite the our model having displacement vector fields as inputs and outputs, we present our inputs and results in the form of positions. For this purpose, we add the displacements in the horizontal and vertical directions to the undeformed positions of the MNIST digit pixels which we assume that lie on a regular grid. The normalized absolute error is computed with respect to the position and not the displacement in each direction.}
\label{fig:displacement_prediction_all}
\end{figure}

\begin{figure}
\centering
\includegraphics[width=1\textwidth]{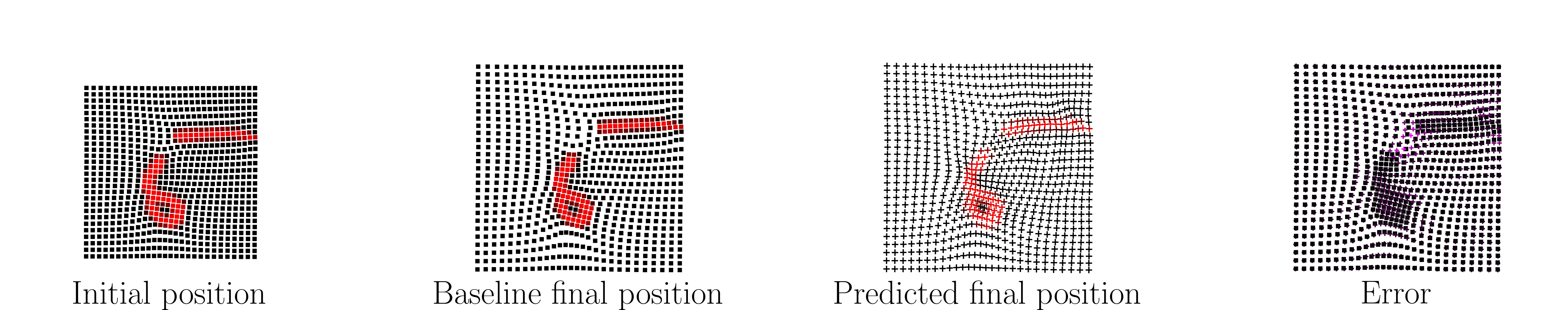}
\caption{Schematic comparison between the predicted and the ground truth final positions for the Mechanical MNIST benchmark: We present the ground truth final and the predicted final positions of the block material together with the point-wise error between them (shown in magenta), as well as the initial position. We present this result in a schematic manner, meaning without some indication of the error magnitude, in order to provide a sense of the deformation of the MNIST pixels under the final displacement.}
\label{fig:displacement_prediction_min}
\end{figure}

We present a visual comparison of the outputs of the trained model against the ground truth solution for three randomly chosen initial conditions from the test data set in Figure \ref{fig:displacement_prediction_all}. Figure \ref{fig:displacement_prediction_min} presents the same comparison for one initial condition to show the change in the pixel position due to the applied displacement, which is not visible in the case where we present multiple solutions at the same time. The error reported in Figure \ref{fig:displacement_prediction_min} illustrates the discrepancy (shown in magenta) between the ground truth and the predicted pixel positions. 

\subsubsection{Shallow Water Equations}
\label{sec:APShallow}

The modeling of the currents in Earth science is often modelled by the Shallow Water equations, which describes the flow below a pressure surface when the horizontal length-scales are much larger than the vertical ones. The system of equations is defined as:
\begin{equation*}
    \begin{split}
        &\frac{\partial \rho}{\partial t} + \frac{\partial (\rho v_1)}{\partial x_1} + \frac{\partial (\rho  v_2)}{\partial x_2} = 0, \\
        &\frac{\partial (\rho  v_1)}{\partial t} + \frac{\partial }{\partial x_1} (\rho  v_1^2 + \frac{1}{2} g \rho^2) + \frac{\partial (\rho v_1 v_2)}{\partial x_2} = 0, \quad \quad t \in (0,1], x \in (0,1)^2 \\
        &\frac{\partial (\rho v_2)}{\partial t} + \frac{\partial (\rho v_1 v_2)}{\partial x_1} + \frac{\partial }{\partial x_2} (\rho v_2^2 + \frac{1}{2}  g \rho^2) = 0,
    \end{split}
    \label{equ:shallow water}
\end{equation*}
where $\rho$ is the total fluid column height, $v_1$ the velocity in the $x_1$-direction, $v_2$ the velocity in the $x_2$-direction, averaged across the vertical column, $\rho$ the fluid density and $g$ the acceleration due to gravity. The above equation can be also written in conservation form:
\begin{equation*}
    \frac{\partial \Vec{U}}{\partial t} + \frac{\partial \Vec{F}}{\partial x_1} + \frac{\partial \Vec{G}}{\partial x_2} = 0,
\end{equation*}
where,
\begin{equation*}
    \Vec{U} = \begin{pmatrix}
 \rho \\
\rho v_1\\
\rho v_2
\end{pmatrix}, \quad \Vec{F} = \begin{pmatrix}
 \rho v_1 \\
\rho v_1^2 + \frac{1}{2} g \rho^2\\
\rho v_1 v_2
\end{pmatrix}, \quad \Vec{G} = \begin{pmatrix}
 \rho v_2 \\
 \rho v_1 v_2 \\
\rho v_2^2 + \frac{1}{2} g \rho^2 
\end{pmatrix}.
\end{equation*}

For a given set of initial conditions, the solution operator $\curlyG$ of \ref{equ:shallow water} maps the initial fluid column height and velocity fields to the fluid column height and velocity fields at later times. Again in this problem the input and the output function domains coincide, therefore $\curlyX = \curlyY$ with $d_x = d_y =3 $ and $d_u = d_s = 3$. The goal is to learn the operator $\curlyG: \curlyC(\curlyX, \mathbb{R}^3) \to \curlyC(\curlyX, \mathbb{R}^3)$. 

We set the boundary conditions by considering a solid, impermeable wall with reflective boundaries:
\begin{equation*}
    v_1 \cdot n_{x_1} + v_2 \cdot n_{x_2} = 0,
\end{equation*}
where $\hat{n} = n_{x_1} \hat{i} + n_{x_2} \hat{j}$ is the unit outward normal of the boundary. We sample the initial conditions by considering a falling droplet of random width, falling from a random height to a random spatial location and zero initial velocities:
\begin{equation*}
    \begin{split}
        &\rho = 1 + h \exp\left(-((x_1 - \xi)^2 + (x_2 -\zeta)^2)/w\right) \\
        &v_1 = v_2 = 0, 
    \end{split}
\end{equation*}
where $\rho$ corresponds to the altitude that the droplet falls from, $w$ the width of the droplet, and $\xi$ and $\zeta$ the coordinates that the droplet falls in time $t=0s$. Because the velocities $v_1,v_2$ are equal to zero in the initial time $t_0=0s$ for all realizations, we choose time $t_0=dt=0.002s$ as the initial time to make the problem more interesting. Therefore, the input functions become
\begin{equation*}
    \begin{split}
        &\rho = 1 + h \exp\left(-((x_1 - \xi)^2 + (x_2 -\zeta)^2)/w\right), \\
        & v_1 = v_1(dt, y_1, y_2), \\
        & v_2 = v_2(dt, y_1, y_2).
    \end{split}
\end{equation*}
We set the random variables $h$, $w$, $\xi$, and $\zeta$ to be distributed according to the uniform distributions
\begin{equation*}
    \begin{split}
        &h= \curlyU(1.5,2.5), \\
        &w = \curlyU(0.002,0.008), \\
        &\xi = \curlyU(0.4,0.6), \\
        &\zeta = \curlyU(0.4,0.6).
    \end{split}
\end{equation*}
The data set is constructed by sampling the initial conditions along a $32 \times 32$ grid and solving the forward problem using a Lax-Friedrichs scheme. This provides us with a solution for a $32 \times 32$ grid which we can use for different operator learning models. Sub-sampling the solution to create the training data set allows us to predict the solution using only a percentage of the available spatial data.

\begin{figure}
\centering
\includegraphics[width=1\textwidth]{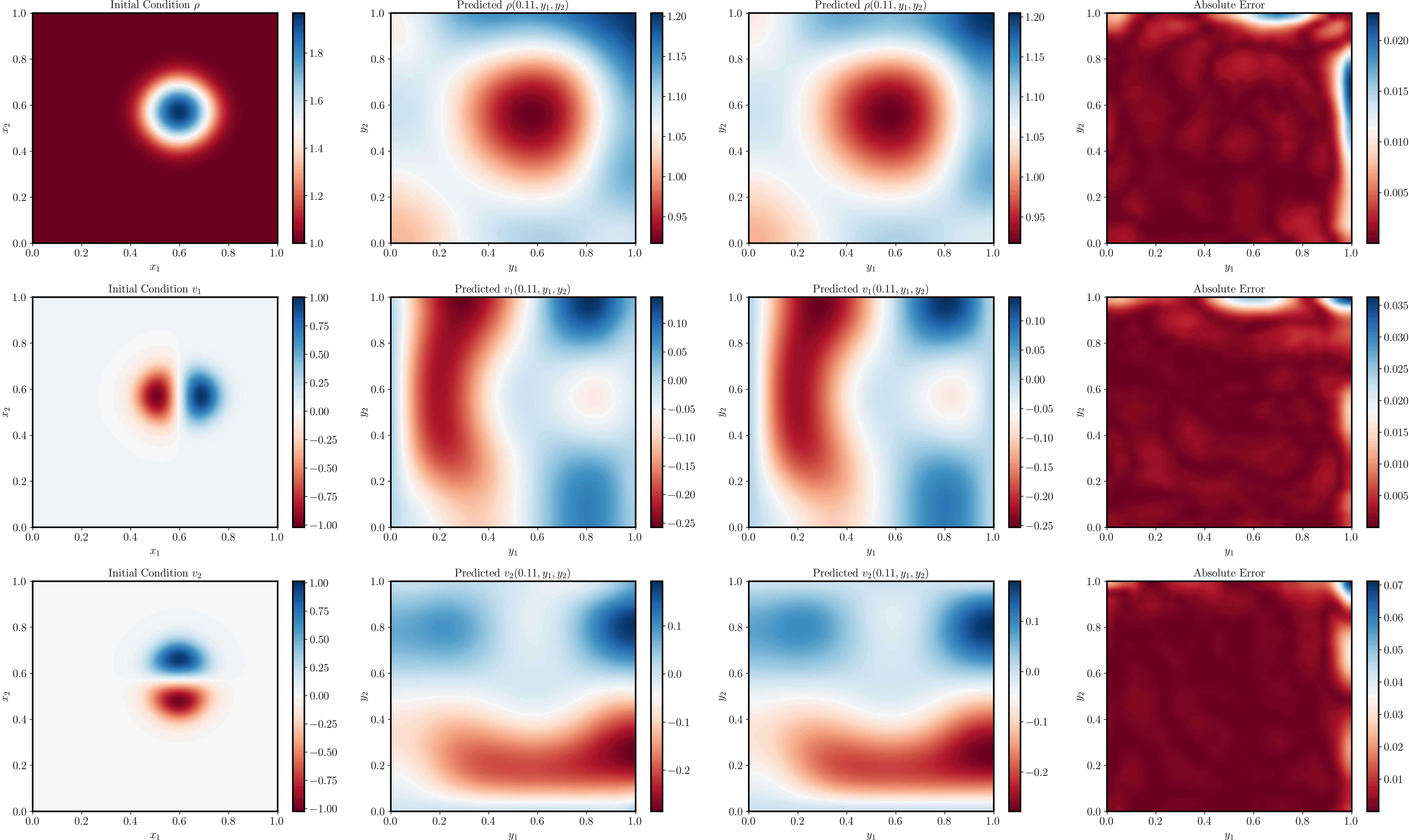}
\caption{Comparison between the predicted and ground truth solution of the Shallow Water Equations benchmark: We present the inputs to the model (initial conditions), the ground truth and the predicted parameters, as well as the absolute error for time instance $t=0.11s$. }
\label{fig:SW_solution_all_t0_testmax_parameter}
\end{figure}

\begin{figure}
\centering
\includegraphics[width=1\textwidth]{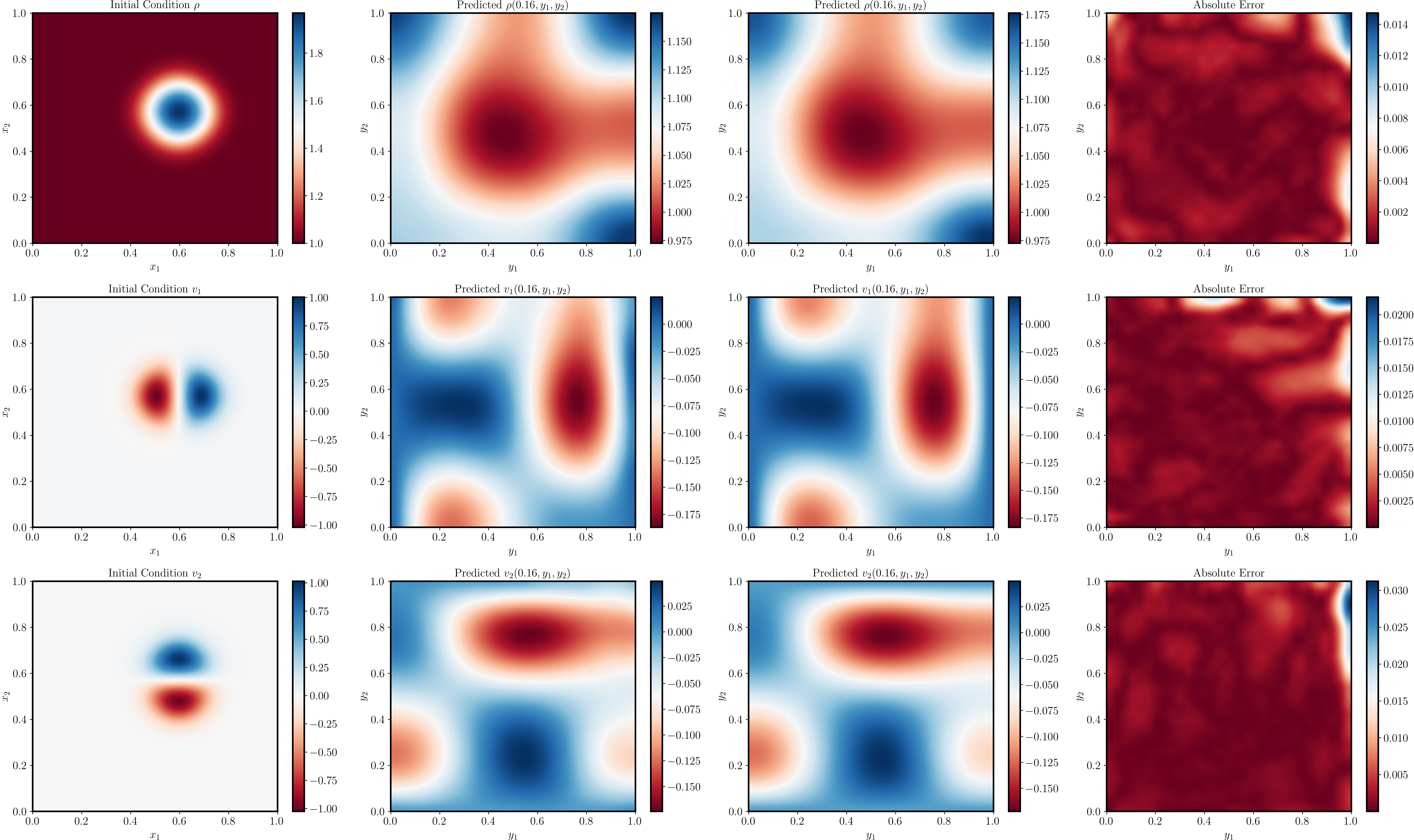}
\caption{Comparison between the predicted and ground truth solution of the Shallow Water Equations benchmark: We present the inputs to the model (initial conditions), the ground truth and the predicted parameters, as well as the absolute error for time instance $t=0.16s$.}
\label{fig:SW_solution_all_t1_testmax_parameter}
\end{figure}

\begin{figure}
\centering
\includegraphics[width=1\textwidth]{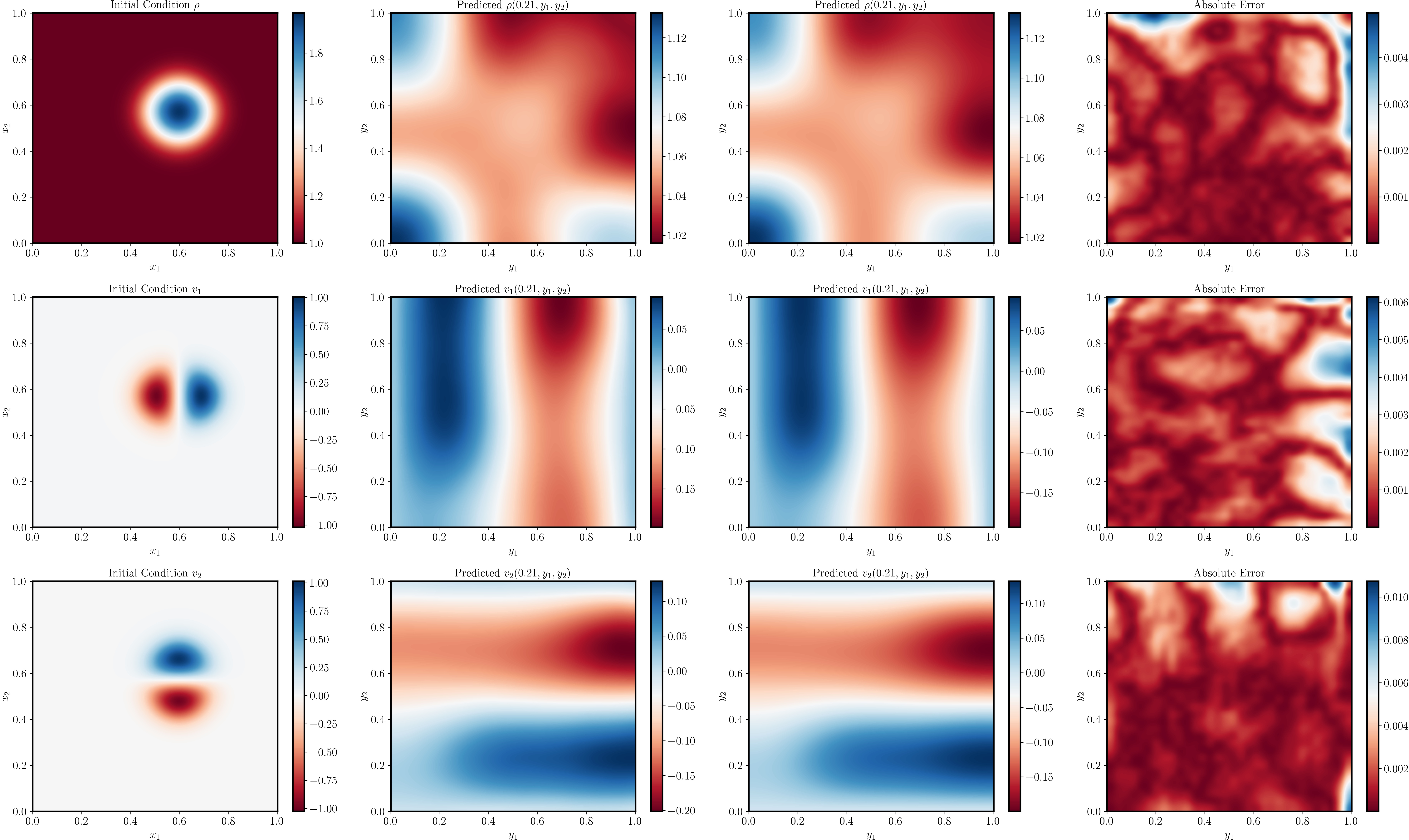}
\caption{Comparison between the predicted and ground truth solution of the Shallow Water Equations benchmark: We present the inputs to the model (initial conditions), the ground truth and the predicted parameters, as well as the absolute error for time instance $t=0.21s$.}
\label{fig:SW_solution_all_t2_testmax_parameter}
\end{figure}

\begin{figure}
\centering
\includegraphics[width=1\textwidth]{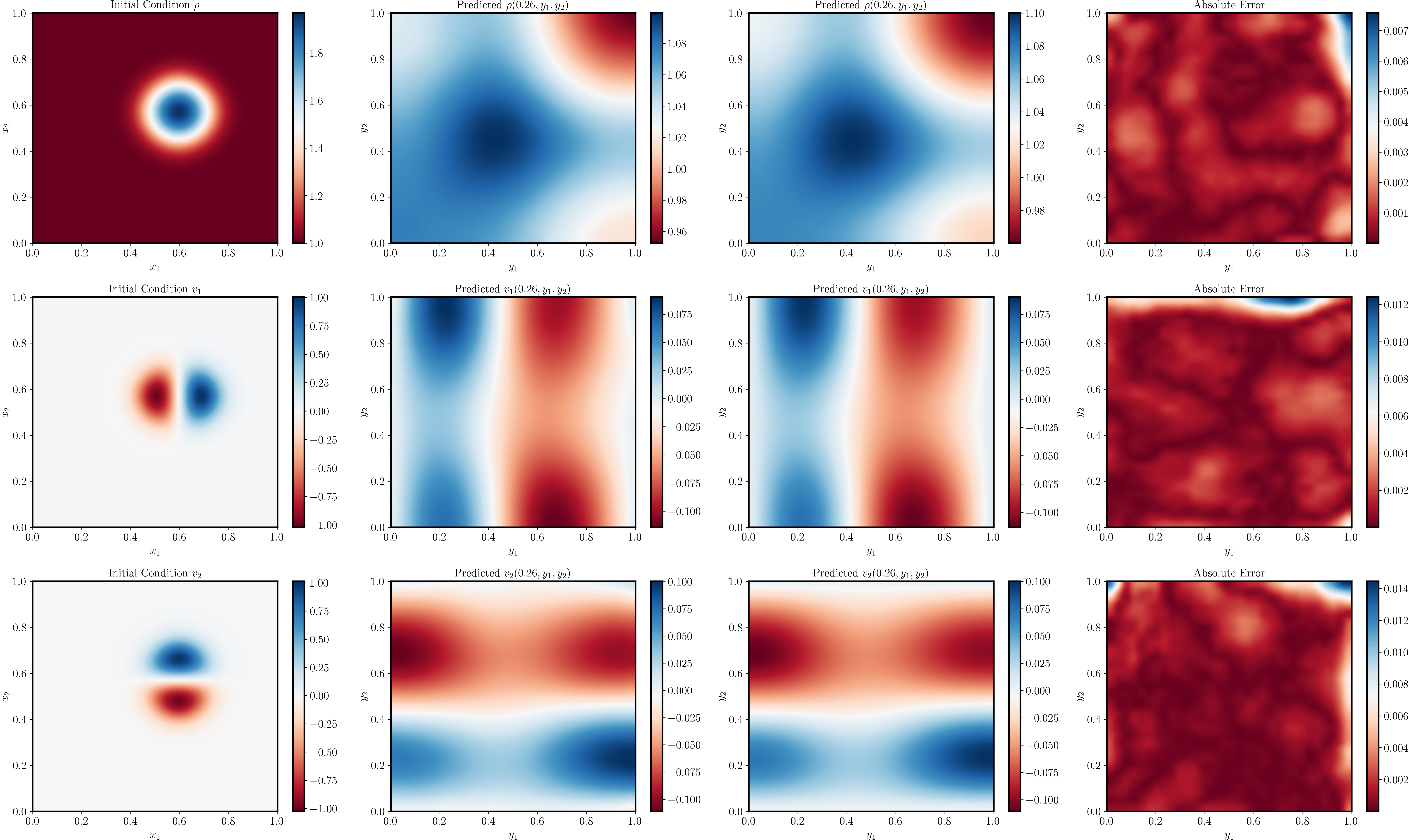}
\caption{Comparison between the predicted and ground truth solution of the Shallow Water Equations benchmark: We present the inputs to the model (initial conditions), the ground truth and the predicted parameters, as well as the absolute error for time instance  $t=0.26s$.}
\label{fig:SW_solution_all_t3_testmax_parameter}
\end{figure}

\begin{figure}
\centering
\includegraphics[width=1\textwidth]{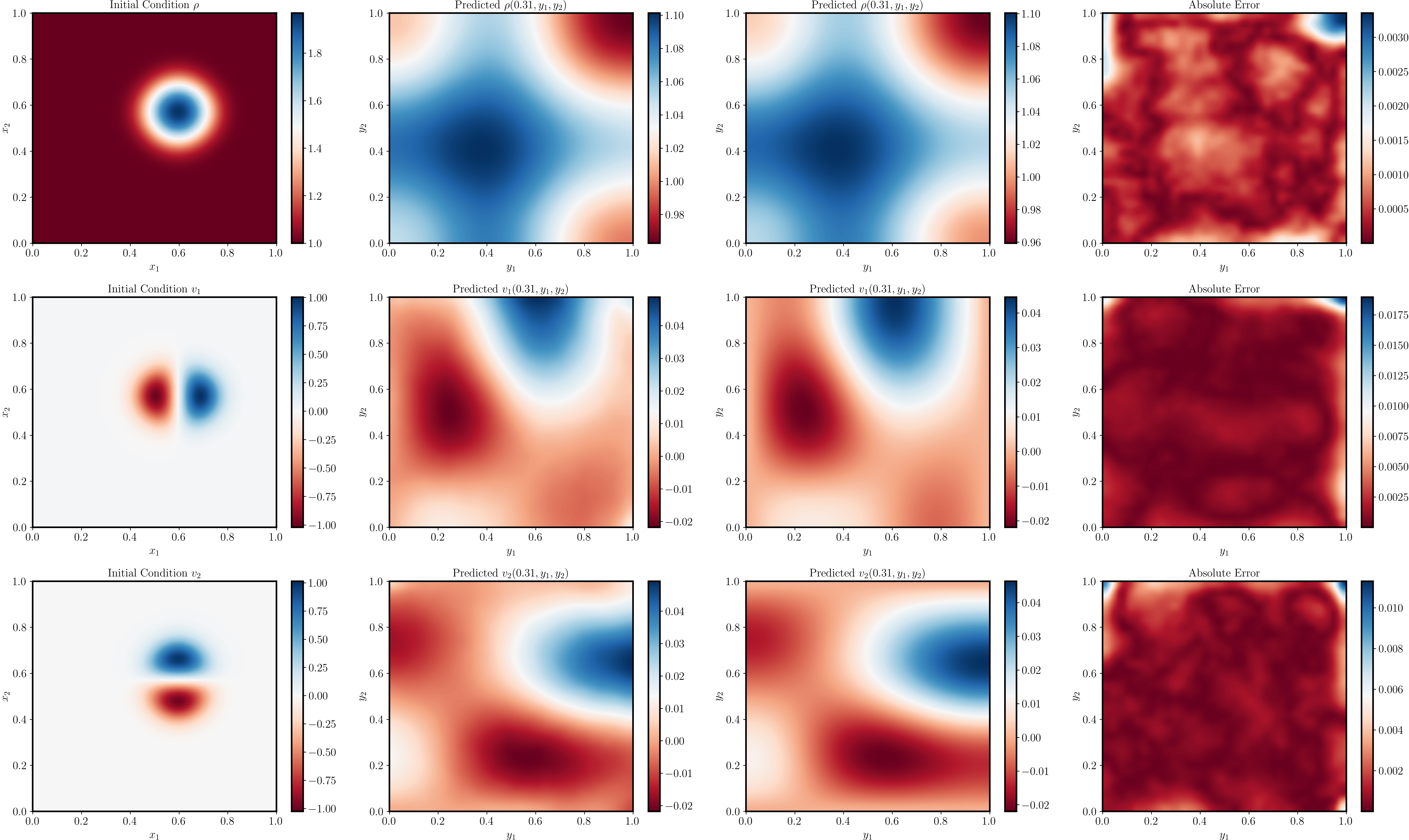}
\caption{Comparison between the predicted and ground truth solution of the Shallow Water Equations benchmark: We present the inputs to the model (initial conditions), the ground truth and the predicted parameters, as well as the absolute error for time instance $t=0.31s$.}
\label{fig:SW_solution_all_t4_testmax_parameter}
\end{figure}

In Figures \ref{fig:SW_solution_all_t0_testmax_parameter}, \ref{fig:SW_solution_all_t1_testmax_parameter}, \ref{fig:SW_solution_all_t2_testmax_parameter}, \ref{fig:SW_solution_all_t3_testmax_parameter}, \ref{fig:SW_solution_all_t4_testmax_parameter} we provide a visual comparison of the outputs of the trained model for 5 time steps, $t = [0.11, 0.16,0.21,0.26,0.31 ]s$, for a randomly chosen initial condition along with the point-wise absolute error plot. We see that our model provides favorable solutions for all time steps for an initial condition not in the train data set. 

\subsubsection{Climate Modeling}
\label{sec:APWeather}
For this example, our aim is to approximate the map between the surface air temperature and surface air pressure. In contrast to the previous examples, here we do not consider a relation between these two fields, for example a partial differential equation or a constitutive law. Therefore, we aim to learn a black-box operator which we then use for making predictions of the pressure using the temperature as an input. Therefore, we consider the map
\begin{equation*}
    T(x) \mapsto P(y),
\end{equation*}
where $x,y \in [-90, 90] \times [0,360] $ for the latitude and longitude. 
For a given day of the year the solution operator maps the surface air temperature to the surface air pressure. For this set-up, the input and output function domains coincide which means $\curlyX = \curlyY$ with $d_x = d_y =2$ and $d_u =d_s = 1$ because we the input and output functions are scalar fields. We can write the map as $\curlyG: \curlyC(\curlyX, \mathbb{R}) \to \curlyC(\curlyX, \mathbb{R})$.

For constructing the training data set, we consider the Physical Sciences Laboratory  meteorological data \cite{kalnay1996ncep}(\url{https://psl.noaa.gov/data/gridded/data.ncep.reanalysis.surface.html}) from the year 2000 to 2005. We consider the different model realizations to be the values of the daily Temperature and Pressure for these 5 years, meaning $N_{train}=1825$ (excluding the days for leap years). We sub-sample the spatial coverage from 2.5 degree latitude $\times$ 2.5 degree longitude global grid ($144 \times 73$) to $72 \times 72$ for creating a regular grid for both the quantities. We consider a test data set consisting of the daily surface air temperature and pressure data from the years 2005 to 2010, meaning $N_{test}=1825$ (excluding leap years), on an $72 \times 72$ grid also.

\begin{figure}
\centering
\includegraphics[width=\textwidth]{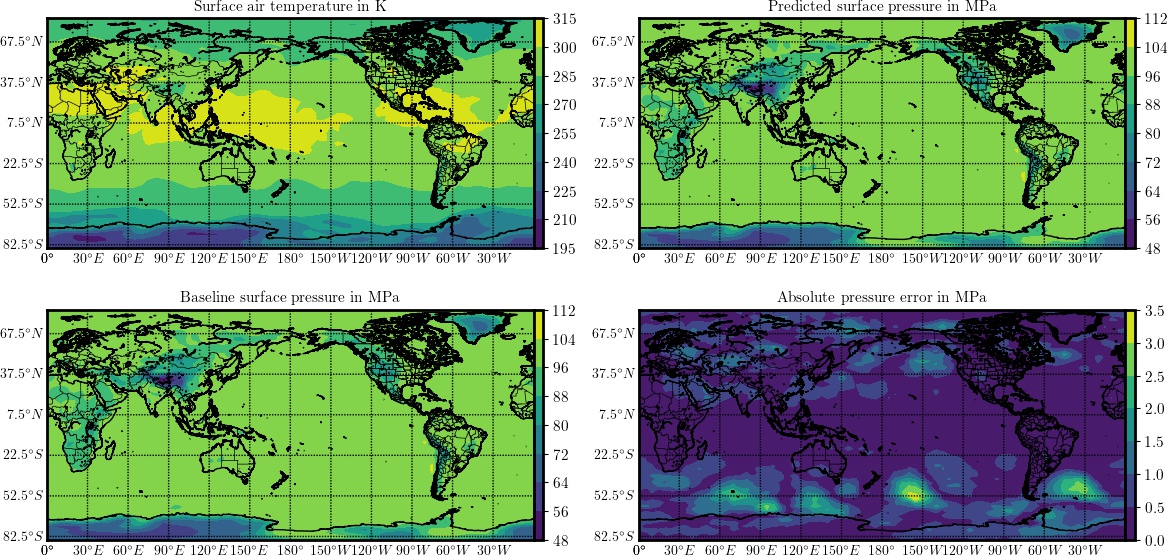}
\caption{Comparison between the full resolution prediction and base line for the climate modeling benchmark: We present the input temperature field, the output prediction and ground truth, as well as the absolute error between our model's prediction and the ground truth solution.}
\label{fig:weather_solution_both_testmin}
\end{figure}

We present the prediction and the ground truth together with the respective input and error  Figure \ref{fig:weather_solution_both_testmin}. The prediction, the ground truth solution and the absolute error are all presented on a $72 \times 72$ grid.

\end{document}